\documentclass{article}

\usepackage{microtype}
\usepackage{graphicx}
\usepackage{color}
\usepackage{booktabs} 
\usepackage{amsmath}
\usepackage{amssymb}
\usepackage{subfig}

\usepackage{colortbl}
\usepackage{xcolor}
\usepackage{pgfplotstable}
\usepackage{booktabs}

\usepackage{colortbl}
\usepackage{pgf}
\usepackage{pgfmath}
\usepackage{xcolor}

\definecolor{high}{RGB}{0,100,0}
\definecolor{low}{RGB}{220,0,0}

\usepackage{hyperref}
\usepackage{multirow}

\def\cca#1{%
    \pgfmathsetmacro\calc{((#1-0)/(100-0))*100}%
    \edef\clrmacro{\noexpand\cellcolor{brown!\calc}}%
    \clrmacro%
    \ifdim \calc pt>12pt\color{black}\fi{#1}%
}

\usepackage[accepted]{icml2025}
\usepackage{adjustbox}

\icmltitlerunning{Gauging Overprecision in LLMs: An Empirical Study}

\begin{document}

\twocolumn[
\icmltitle{Gauging Overprecision in LLMs: An Empirical Study}




\begin{icmlauthorlist}
\icmlauthor{Adil Bahaj}{to}
\icmlauthor{Hamed Rahimi}{goo}
\icmlauthor{Mohamed Chetouani}{goo}
\icmlauthor{Mounir Ghogho}{to}
\end{icmlauthorlist}

\icmlaffiliation{to}{COLCOM, University Mohammed VI Polytechnic, Rabat Morocco}
\icmlaffiliation{goo}{ISIR, Sorbonne University, Paris, France}

\icmlcorrespondingauthor{Adil Bahaj}{adil.bahaj@um6p.ma}

\icmlkeywords{Large language models, overconfidence, overprecision}

\vskip 0.3in
]



\printAffiliationsAndNotice{}  

\begin{abstract}

Recently, overconfidence in large language models (LLMs) has garnered considerable attention due to its fundamental importance in quantifying the trustworthiness of LLM generation. However, existing approaches prompt the \textit{black box LLMs} to produce their confidence (\textit{verbalized confidence}), which can be subject to many biases and hallucinations. Inspired by a different aspect of overconfidence in cognitive science called \textit{overprecision}, we designed a framework for its study in black box LLMs. This framework contains three main phases: 1) generation, 2) refinement and 3) evaluation. In the generation phase we prompt the LLM to generate answers to numerical questions in the form of intervals with a certain level of confidence. This confidence level is imposed in the prompt and not required for the LLM to generate as in previous approaches. We use various prompting techniques and use the same prompt multiple times to gauge the effects of randomness in the generation process. In the refinement phase, answers from the previous phase are refined to generate better answers. The LLM answers are evaluated and studied in the evaluation phase to understand its internal workings. This study allowed us to gain various insights into LLM overprecision: 1) LLMs are highly uncalibrated for numerical tasks 2) { there is no correlation between the length of the interval and the imposed confidence level, which can be symptomatic of a a) lack of understanding of the concept of confidence or b) inability to adjust self-confidence by following instructions}, { 3)} LLM numerical precision differs depending on the task, scale of answer and prompting technique { 4) Refinement of answers doesn't improve precision in most cases}. We believe this study offers new perspectives on LLM overconfidence and serves as a strong baseline for overprecision in LLMs.
\end{abstract}
\section{Introduction}
Overconfidence is a cognitive bias that affects human decision-making, characterized by a level of confidence that exceeds what is justified by reality. In cognitive science, overconfidence has been studied across three distinct dimensions \cite{moore2017individual, moore2017three}: (1) Overestimation, (2) Overplacement, and (3) Overprecision. Overestimation involves an inflated perception of one's abilities or performance relative to their actual level. Overplacement refers to an exaggerated belief in one's superiority over others. Overprecision is defined as unwarranted certainty in the accuracy of one's knowledge or beliefs. Among these dimensions, overprecision is considered the most robust \cite{moore2015overprecision, moore2015wide}, as it consistently lacks contradictory findings across different studies, unlike the other aspects.

\begin{figure*}[h]
    \centering
    \includegraphics[scale=0.5]{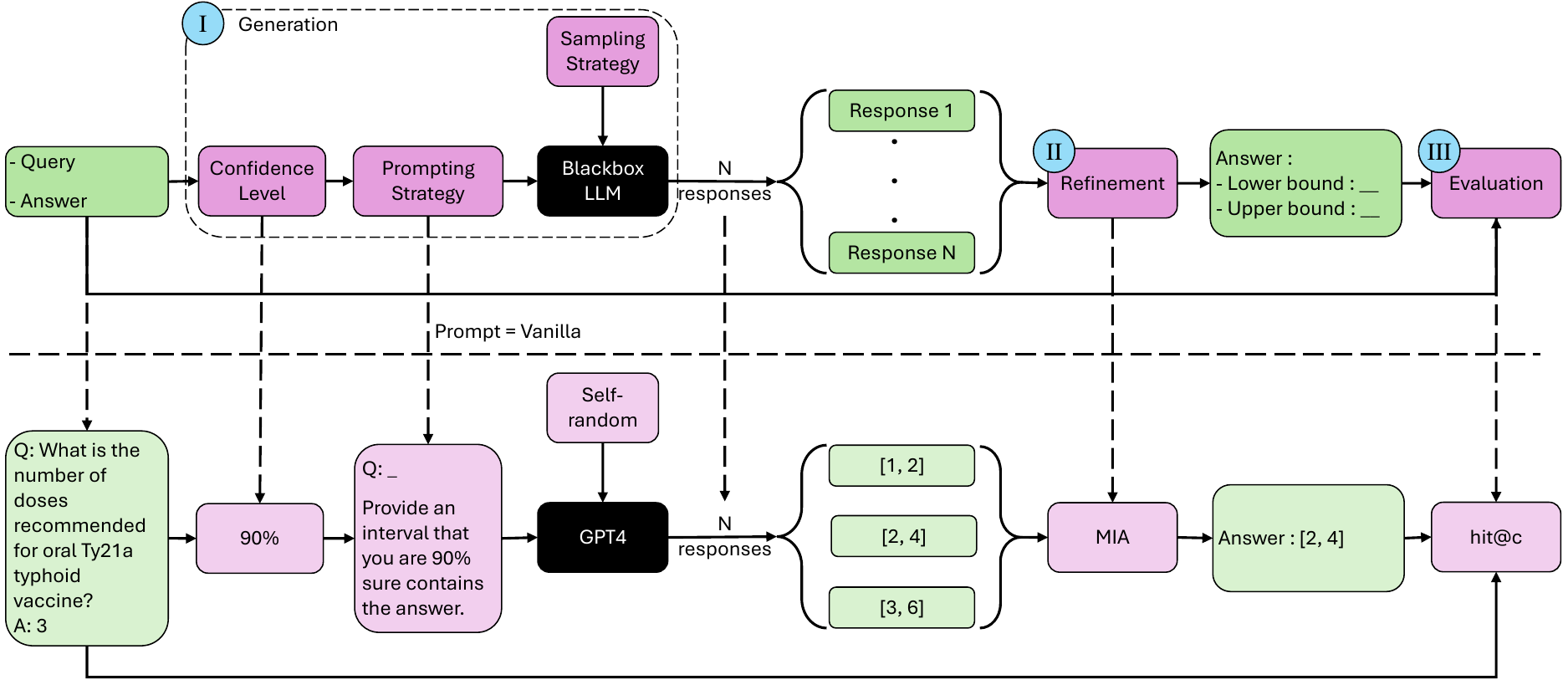}
    \caption{An outline of the precision elicitation framework and an example. Given an input question, a confidence level is first specified, a prompt strategy is then chosen, and the confidence level is integrated into the prompt. Next, the sampling strategy and the number of samples are determined to control the amount and diversity of outputs of the same prompt. After that, an \textit{aggregator} combines the different answers to produce the most likely answer.}
    \label{fig:framework}
\end{figure*}

Our study addresses a critical gap in overconfidence research by focusing on overprecision in black-box LLMs \cite{bahaj2024asthmabot, achiam2023gpt}. Our key contributions are: (1) constructing datasets specifically designed to evaluate overprecision, (2) designing an experimental protocol to systematically investigate overprecision in LLMs, and (3) conducting a comparative analysis to study the impact of different techniques. The proposed framework is structured into three phases: generation, refinement, and evaluation. In the generation phase, the LLM generates numerical intervals at specified confidence levels using multiple prompts to account for randomness. This phase leverages the inherent instruction-following capabilities of LLMs to improve overconfidence quantification. In the refinement phase, the generated responses are improved for greater reliability through two strategies: (1) aggregation, where intervals are merged to enhance accuracy, and (2) self-refinement, where the LLM evaluates and refines its own responses. Finally, the evaluation phase measures the LLM's performance across tasks using cognitive science-inspired metrics, enabling a comprehensive analysis of its behavior. An overview of this framework is presented in Figure \ref{fig:framework}.

This study highlights key findings: (1) LLMs are poorly calibrated for numerical answers; (2) { there is no correlation between the length of the interval and the imposed confidence level, which can be symptomatic of a a) lack of understanding of the concept of confidence or b) inability to adjust self-confidence by following instructions}; (3) numerical precision depends on the task, answer scale, and prompts; and (4) while refinement strategies can improve precision, most offer limited gains. Surprisingly, self-refinement significantly reduces performance, contrasting with prior cognitive science and LLM studies \cite{haran2010simple, xiongcan}.

\section{Related Work}
    \subsection{Overconfidence in Humans}
        \label{sec:overconf_human}
        Overconfidence is an unwarranted certainty in one’s knowledge or abilities \cite{kruger1999unskilled}, often associated with negative consequences in fields such as medicine \cite{al2024overconfidence, seidel2024interaction}, politics \cite{ortoleva2015overconfidence}, and finance \cite{grevzo2021overconfidence}. It is traditionally studied across three dimensions: overestimation, overplacement, and overprecision \cite{moore2017three, moore2017individual}.
        Overestimation refers to an inflated perception of one’s abilities and is commonly assessed through item-confidence judgments, where participants respond to general knowledge questions and rate their confidence levels \cite{harvey1997confidence}. Overplacement explores the "better-than-average" effect, where individuals mistakenly believe they are superior to others, often resulting in the majority of participants rating themselves as above average \cite{beer2010neural}. Overprecision captures unwarranted certainty in the accuracy of one’s estimates and is typically measured by asking participants to define narrow confidence intervals around their best guesses \cite{alpert1982progress}.
        Among these dimensions, overprecision is the most robust, consistently demonstrated across studies, whereas overestimation and overplacement often produce inconsistent findings \cite{moore2015overprecision, moore2015wide}. This work focuses on the study, measurement, and quantification of overprecision in LLMs.

    \subsection{Overconfidence in LLMs}
        { 
        Overconfidence has been studied extensively in the literature \cite{geng2024survey}. Approaches for overconfidence estimation in LLMs can be categorized depending on the kinds of models they are applied to: a) white-box, b) black-box. White-box approaches have access to the internal workings and calculation of an LLM, which they use to estimate overconfidence \cite{huang2024calibrating, duanshifting}. However, black-box approaches lack any access to the internal processing of LLMs, which they surpass by devising prompting techniques \cite{manakul-etal-2023-selfcheckgpt, mielke2022reducing, xiongcan} or surrogate models \cite{shrivastava2023llamas}. This work belongs to the black-box paradigm.}
        {  \subsubsection{Overconfidence in Black Box LLMs}}
        Previous approaches to studying overconfidence have primarily focused on the overestimation aspect \cite{wen2024mitigating, xiongcan, geng2024survey}. These studies typically rely on eliciting an LLM's confidence in its answers, which presents significant limitations, as LLMs are generally not trained to introspect or reflect on their internal knowledge. Furthermore, LLMs are not optimized for self-reflection but are designed to follow instructions. Additionally, LLM outputs are prone to hallucinations, a problem that is exacerbated when confidence is elicited for inherently subjective measures like self-confidence, raising concerns about the validity of many confidence elicitation methods. To address these limitations, this work proposes a novel approach in which a confidence level is explicitly imposed within the prompt, requiring the LLM to adhere to this confidence level when answering questions. This method leverages the natural instruction-following capabilities of LLMs. Moreover, the study focuses on numerical answers rather than categorical ones, enabling a more nuanced examination of LLM confidence while avoiding biases commonly associated with categorical responses \cite{sumita2024cognitive}. {  Recently, \cite{groot2024overconfidence} designed various prompts for regression tasks for confidence estimation in vision LLMs. This approach for numerical reasoning differs from ours in many aspects. First, the authors employed a confidence verbalisation approach similar to that described in \cite{xiongcan}. Second, the authors tried to estimate confidence in visual perception, not knowledge. This can be considered a sub-task of confidence in knowledge since the vision LLM is provided with contextual information is only tasked to "see", not "remember", and "reason".}

\section{Overprecision in Black Box LLMs}
    Let ${(q_{i}, a_{i})}_{i}$ represent a set of questions and their corresponding answers, where $q{i}$ is a textual question, and $a_{i} \in \mathbb{R}$ is its numerical answer. This work proposes a framework for studying overprecision in LLMs, consisting of three phases: (a) generation, (b) refinement, and (c) evaluation.
    The generation phase involves generating (i.e., predicting) an answer for each question using an existing LLM. The refinement phase takes the answers produced during the generation phase and applies various techniques to rectify and improve these answers. Finally, the evaluation phase analyzes the answers from the previous phases to assess the precision and confidence of the LLM. The details of each phase and its corresponding steps are presented in the following sections.
    
\subsection{Generation}
    The objective of the generation step is to produce answers using an LLM. The generation process consists of two main components: (a) {\em prompting strategy} and (b) {\em sampling strategy}. The prompting strategy involves integrating the question into a confidence-parametrized prompt composed of various parts. This prompt, or its variants, is then provided to the LLM multiple times, following a specific sampling strategy.
    Formally, this phase is responsible for constructing a prompt $\mathbf{p}_{c}(q)$ parameterized by a confidence level $c$. This prompt is fed into the LLM to generate a lower bound $x$ and an upper bound $y$, defining the interval within which the answer to the question $q$ should fall:
    \begin{equation}
        (x, y)=\text{LLM}(\mathbf{p}_{c}(q))
    \end{equation}
\subsubsection{Prompting Strategy}
    Let $\mathbf{p}_{c}$ represent a prompt parameterized by a confidence level $c$. This prompt includes a series of instructions that the LLM must follow to answer the question. These instructions can be divided into distinct sets. Formally, $\mathbf{p}_{c}$ can be expressed as:
    \begin{equation}
        \label{eq:vanilla_prompt}
        \mathbf{p}_{c}(q_{i})=[\text{GEN}, \text{CONF}_{c}, \text{CONFK}, \text{FORM}, \text{QUES}(q_{i})]
    \end{equation}
   where $[.]$ denotes text concatenation. Table \ref{tab:inst_prompt} provides further details on the formulation and purpose of each instruction set. The initial prompt employs a vanilla prompting strategy. An alternative experimental variant utilizes the chain of thought (CoT) prompting strategy and is formulated as follows:
   \begin{equation}
        \label{eq:cot_prompt}
        \mathbf{p}_{c}(q_{i})=[\text{GEN}, \text{CONF}_{c}, \text{CONFK}, \text{FORM}, \text{CoT}, \text{QUES}(q_{i})]
    \end{equation}
    The formulation of $\text{CoT}$ is in table \ref{tab:inst_prompt}.
   \begin{table*}[]
       \centering
       \scalebox{0.8}{
       \begin{tabular}{c|p{16em}|p{16em}}
            Instruction& Text & Objective\\
            \hline
            $\text{GEN}$&"please follow these instructions to ..."&General instructions that the LLM should follow\\
            $\text{CONF}_{c}$& "Please give us two numbers: a ‘lower bound’ and an ‘upper bound’... you should be { $c$\%} sure that the answer falls between the lower and upper bounds" & Instructing the LLM on the level of confidence that it should have in its answer.\\
            $\text{CONFK}$&"The more unsure you are in your response ..."&Giving the LLM general knowledge about confidence\\
            $\text{FORM}$&"your answer should have the following format ..."& Formating instructions that facilitate the parsing of the LLM output\\
            $\text{CoT}$& "give your step-by-step reasoning for why..."&Chain of Thought instructions for better reasoning.\\
            $\text{HINT}$&"I read in a book that the right answer is: { [lower\_bound, upper\_bound]}..."&Misleading hint given to the LLM to gouge its true confidence.\\
            $\text{QUES}(q_{i})$&"Question: { [$q_{i}$]}"&The question that the LLM should answer.\\
       \end{tabular}}
       \caption{Sets of instructions that are used in the prompts. 'instruction' represents the abbreviation used in the paper for a particular set of instructions. 'Text' is the instruction text. 'objective' is the purpose of having that set of instructions.}
       \label{tab:inst_prompt}
   \end{table*}
\subsubsection{Sampling Strategy}
    We employed the following sampling strategies: (a) {\em self-random} and (b) {\em misleading}. The self-random sampling strategy involves prompting the LLM multiple times to leverage the inherent randomness of the generation process. The prompts defined in Eqs. \ref{eq:vanilla_prompt} and \ref{eq:cot_prompt} are repeatedly fed to the LLM to obtain randomly sampled answers.
    
    The misleading strategy aims to deceive the LLM into providing incorrect answers by introducing a random answer, e.g., “I read in a textbook that the answer is ...”. This approach is designed to introduce doubt into the LLM's reasoning process to assess its true confidence. These misleading hints are incorporated into the prompts, modifying them such that the vanilla prompt in Eq. \ref{eq:vanilla_prompt} becomes:
    \begin{equation}
        \label{eq:vanilla_prompt}
        \mathbf{p}_{c}(q_{i})=[\text{GEN}, \text{CONF}_{c}, \text{CONFK}, \text{FORM}, \text{HINT}, \text{QUES}(q_{i})]
    \end{equation}
    and the CoT prompt in eq. \ref{eq:cot_prompt} becomes
    \begin{equation}
        \label{eq:cot_prompt}
        \mathbf{p}_{c}(q_{i})=\begin{split}
            [\text{GEN}, \text{CONF}_{c}, &\text{CONFK}, \text{FORM},\\ &\text{CoT}, \text{HINT}, \text{QUES}(q_{i})]
        \end{split}
    \end{equation}
\subsection{Refinement}

    We investigate two refinement strategies: (a) {\em Aggregation} and (b) {\em Self-refinement}. Aggregation involves combining multiple output intervals to generate an interval that is most likely to contain the correct answer. While aggregation methods are well-studied for categorical outputs, limited work exists for numerical outputs. To bridge this gap, we propose several novel aggregation techniques. Self-refinement utilizes the LLM's own outputs by feeding them back into the model, allowing it to evaluate the responses, select the most probable answer, and suggest improvements. This approach is inspired by cognitive science research on overprecision, which demonstrates that access to peer responses can enhance precision.
\subsubsection{Aggregation strategies}
    Let ${[x_{i}, y_{i}]}_{i=1}^{N}$ represent a set of $N$ intervals obtained by prompting the LLM $N$ times using variants of the previously discussed prompts. Let $c_i$ denote the confidence level imposed on the LLM in the prompt to generate the $i$th answer. Interval aggregation combines the upper and lower bounds of these output intervals to produce an aggregated interval. Formally, this strategy can be defined as follows:
            \begin{equation}
            X =\frac{\sum_{i=1}^{N}w_{i}x_{i}}{\sum_{j=1}^{N}w_{i}}, \qquad
            Y =\frac{\sum_{i=1}^{N}w_{i}y_{i}}{\sum_{j=1}^{N}w_{i}}
    \end{equation}

    where $X$ and $Y$ are the lower and upper bounds of the aggregated interval, respectively, and $w_{i}$ is a weight that determines the contribution of the $i$th interval to the overall aggregation. The values of the $w_{i}$'s are determined based on various weighting schemes. In this study, we utilized the following:
    \begin{itemize}
        \item Mean interval aggregation (MIA): This strategy gives each interval equal weighting as follows: $w_{i} = 1,\forall i$.
        \item Length weighted aggregation (LWA): This strategy weighs longer intervals more than smaller intervals as follows: $w_{i}=d_{i}, \forall i$, where $d_{i}=y_{i}-x_{i},\forall i$.
        
        \item Inverse length weighted aggregation (iLWA): This strategy weighs shorter intervals more than longer intervals as follows: $w_{i}=\bar{d}_{i},\forall$, where $\bar{d}_{i} = \frac{1}{y_{i}-x_{i}},\forall i$.
        
        \item Confidence weighted aggregation (CWA): in cases where the same query is prompted at different confidence levels, confidence intervals can be used to weigh the intervals as follows: $w_{i}=c_{i},\forall i$.
        
    \end{itemize}
    In addition to the previous schemes, we also experiment with the union of intervals (Union), which can be presented formally as follows:
    \begin{equation}
            X=\min(\{x_i\}_i), \qquad
            Y=\max(\{y_i\}_i)
    \end{equation}

\subsubsection{Self-refinement}
    For a set of $N$ responses and their corresponding confidence levels, $A = {[x_{i}, y_{i}, c_{i}]}_{i=1}^{N}$, obtained during the generation step for a question $q$, self-refinement involves improving the LLM's responses by prompting it to evaluate the initial answers, select the most probable one, and propose an enhanced response. This process takes into account the confidence levels associated with each answer generated in the initial step. Formally, this process can be expressed as follows:
    \begin{equation}
        \label{eq:selfref}
        \text{LLM}(\mathbf{p}^{\text{refine}}({[x_{i}, y_{i}, c_{i}]}_{i=1}^{N}, q, e))=\begin{cases}
            (X^{\rm old}, Y^{\rm old})\\
            (X^{\rm new}, Y^{\rm new})
        \end{cases}
    \end{equation}
    where $X^{\rm old} \in \{x_i\}_i$ and $Y^{\rm old} \in \{y_i\}_i$ are bounds from the existing list of proposed bounds within which the potential answer may lie; $X^{\rm new}$ and $Y^{\rm new}$ represent the new lower and upper bounds, respectively, generated by the LLM based on the potential answers and their associated confidence levels; and $e$ denotes the number of elements sampled from $A$. Table \ref{tab:selfrefineprompt} provides a summary of the formulation of the self-refine prompt.
    \begin{table}[]
        \centering
        \begin{tabular}{|p{20em}|}
            \hline
             prompt $\mathbf{p}^{\text{refine}}({[x_{i}, y_{i}, c_{i}]}_{i=1}^{N}, q, e))$  \\
             \hline
            - Context: A group of people were given a question ...\\
            - Instructions:\\
              - Analyse the question, the answers to the question and their corresponding confidence level.\\
              - Determine the most likely ...\\
              - give your reasoning ...\\
              - Your output should have the following format ...:\\
              \{
                "chosen\_answer":[lower\_bound, upper\_bound],
                "chosen\_reason":,
                "proposed\_answer":[lower\_bound, upper\_bound],
                "proposed\_reason":
              \}\\
            - Question: { $q$}\\
            - Possible Answers:\\
            { 
                $
                e \text{ examples}= \begin{cases}
                    x_i| y_i| c_i\\
                    \cdots\\
                    x_j| y_j| c_j
                \end{cases}$                
            }\\
            \hline
        \end{tabular}
        \caption{Self-refinement prompt. The prompt takes as inputs a question $q$ and a set of $e$ potential answers from the generation phase.}
        \label{tab:selfrefineprompt}
    \end{table}
\subsection{Evaluation}
    We evaluate the LLM on two primary tasks: (a) precision calibration and (b) confidence understanding. Let $\hat{A}^c = {(q_i, a_i, [x_{i}^c, y_{i}^c])}_i$ represent a set of questions $q_i$ with their corresponding ground truth answers $a_i$ and the LLM-generated intervals $[x{i}^c, y_{i}^c]$ at a confidence level $c$, obtained using a variation of the previously discussed prompting techniques.
    In line with existing literature on overprecision in cognitive science \cite{soll2004overconfidence, moore2015wide}, we use the hit metric, which calculates the percentage of instances where the ground truth answers fall within the generated intervals. Formally, this can be expressed as follows:
    \begin{equation}
        \text{hit}@c\%=\frac{1}{|\hat{A}^c|} \sum_{i=1}^{|\hat{A}^c|}I(a_i\in [x_{i}^c, y_{i}^c])
    \end{equation}
   where $I$ is the indicator function, defined as $I(\text{cond}) = 1$ if the condition $\text{cond}$ is satisfied, and $I(\text{cond}) = 0$ otherwise. Additionally, we compute Pearson's correlation coefficient \cite{sedgwick2012pearson} between the confidence levels and the lengths of the intervals to assess the LLM's awareness of its own self-confidence \cite{moore2008trouble}.

{ 
\subsection{Motivation}
    Our methodology focuses on numerical reasoning for various reasons. First, this focus mirrors the studies of overprecision in cognitive science, which is a more consistently measured aspect of overconfidence relative to overestimation and overclaiming (section \ref{sec:overconf_human}). Second, we hypothesise that focusing on numerical outputs instead of categorical or mixed outputs gives a better measure for a model's general overconfidence since it avoids various cognitive biases related to language, such as positivity bias \cite{sumita2024cognitive}. Third, as opposed to previous works \cite{xiongcan} that focused on direct question/answer format and multi-choice questions (MCQ) format, we only focus on the direct question/answer format to avoid the different biases that LLMs exhibit in MCQs, such as order bias and authoring bias \cite{sumita2024cognitive, zhenglarge}.

}
\section{Experimental Setup}
\paragraph{Datasets}
    We utilized the following datasets: FinQA \cite{chen2021finqa}, MedMCQA \cite{pal2022medmcqa}, MedQA \cite{jin2021disease}, and MMLU \cite{hendrycksmeasuring}. FinQA is designed for numerical reasoning over financial data. MedMCQA and MedQA are datasets consisting of medical multiple-choice questions (MCQs). MMLU is a versatile dataset that spans multiple domains, tasks, and topics. These datasets were selected to capture a range of numerical reasoning complexities. While MMLU focuses on general knowledge, FinQA and the medical datasets require more domain-specific expertise. FinQA, in particular, presents an additional level of difficulty as it involves reasoning directly from specialized financial reports of companies.
    
\paragraph{Data Processing}    
    These datasets were filtered to extract questions with numerical answers that do not include units of measure, currency symbols, or any other strings conveying additional information about the number. Multiple-choice question (MCQ) data was converted to direct answer format, ensuring that each question has a single answer without any options. Due to the limited number of numerical answers in the test splits of these datasets, we sampled questions from all splits during the process. Additionally, MedMCQA and MedQA were combined into a single dataset referred to as "Medical." Table \ref{tab:data_info} outlines the key characteristics of these datasets.
    \begin{table}[]
        \centering
        \scalebox{0.8}{
        \begin{tabular}{c|c|c|c|c}
            dataset& \#examples & avg-a & min-a & max-a \\
            \hline
            FinQA & 3262&1.109e+08 &-2.094e+09 &8.096e+10\\
            Medical & 2058& 4.033e+03&-1.000e+02&6.123e+06\\
            MMLU &1606&1.222e+10 & -1.280e+02 & 9.789e+12\\
        \end{tabular}}
        \caption{Summary statistics of the different datasets. "\#examples" is the number of question/answer pairs in the dataset. avg-a, min-a and max-a are the mean, minimum and maximum of the ground truth answers in the datasets.}
        \label{tab:data_info}
    \end{table}
\paragraph{Models}
    We focused on widely adopted black-box LLM models with established reliability, including GPT-3.5-turbo \cite{schulman2022chatgpt} and GPT-4o-Mini \cite{achiam2023gpt}.
\subsection{Protocol}

    \paragraph{Phase 1 (Generation)} Each question in the dataset is paired with a specific prompting strategy, sampling strategy, and confidence level ([60\%, 70\%, 80\%, 90\%, 95\%]). These combinations are evaluated on an LLM over five trials to account for randomness. Each trial produces an interval with upper and lower bounds for the predicted answer.

    \paragraph{Phase 2 (Refinement)} Answers generated in the first phase are refined using either aggregation or self-refinement strategies. For each question-answer pair, responses are sampled and processed through a refinement function to produce a new interval. To ensure cost efficiency, a single model is utilized throughout this phase. Two settings are considered: (1) Mixed confidence, where responses are sampled randomly across different confidence levels, and (2) Single confidence, where responses are sampled randomly within a specific confidence level.

   For each combination, a single trial is randomly sampled, and evaluation metrics are computed over 10 iterations. Both the mean and standard deviation are reported. Due to budget constraints, multiple prompts were not feasible for self-refinement; thus, a single trial per question-answer pair was used. This approach relies on prior experiments (i.e., the generation phase) to assume consistency in the results.
    
\section{Evaluation and Analysis}
\label{sec:res_eval}

\begin{table*}[!h]
\centering
\scalebox{0.8}{
    \setlength{\tabcolsep}{1.7pt}
    
\begin{tabular}{lll|r|r|r|r|r|r|r|r|r|r|r|r|r|r}
\toprule
 &  &  & \multicolumn{2}{c|}{hit@95\%} & \multicolumn{2}{c|}{hit@90\%} & \multicolumn{2}{c|}{hit@80\%} & \multicolumn{2}{c|}{hit@70\%} & \multicolumn{2}{c|}{hit@60\%} & \multicolumn{2}{c|}{hit-avg} & \multicolumn{2}{c}{corr} \\
dataset & model & P.S. & mean & std & mean & std & mean & std & mean & std & mean & std & mean & std & mean & std \\
\midrule
\multirow[t]{4}{*}{FinQA} & \multirow[t]{2}{*}{gpt-3.5-turbo} & vanilla & \cca{6.16} & 0.24 & \cca{5.50} & 0.23 & \cca{6.47} & 0.30 & \cca{6.79} & 0.20 & \cca{7.42} & 0.28 & \cca{6.47} & 0.09 & -0.0089 & 0.0070 \\
 &  & CoT & \cca{7.04} & 0.25 & \cca{7.16} & 0.36 & \cca{7.33} & 0.49 & \cca{7.35} & 0.34 & \cca{7.55} & 0.21 & \cca{7.29} & 0.17 & 0.0034 & 0.0143 \\
 & \multirow[t]{2}{*}{gpt-4o-mini} & vanilla & \cca{21.14} & 0.35 & \cca{18.95} & 0.41 & \cca{18.25} & 0.36 & \cca{16.05} & 0.43 & \cca{17.04} & 0.45 & \cca{18.29} & 0.20 & -0.0019 & 0.0038 \\
 & & CoT & \textbf{\cca{21.54}} & 0.41 & \textbf{\cca{20.29}} & 0.51 & \textbf{\cca{20.32}} & 0.43 & \textbf{\cca{19.05}} & 0.41 & \textbf{\cca{19.75}} & 0.46 & \textbf{\cca{20.19}} & 0.12 & -0.0006 & 0.0089 \\
 \hline
\multirow[t]{4}{*}{Medical} & \multirow[t]{2}{*}{gpt-3.5-turbo} & vanilla & \cca{48.28} & 0.59 & \cca{47.71} & 0.59 & \cca{48.85} & 0.84 & \cca{47.26} & 0.55 & \cca{49.42} & 0.72 & \cca{48.31} & 0.25 & -0.0051 & 0.0089 \\
 & & CoT & \cca{48.48} & 0.60 & \cca{47.79} & 0.74 & \cca{49.60} & 0.99 & \cca{49.46} & 1.11 & \cca{48.68} & 0.89 & \cca{48.80} & 0.38 & -0.0004 & 0.0094 \\
 & \multirow[t]{2}{*}{gpt-4o-mini} & vanilla & \cca{60.31} & 0.55 & \cca{60.41} & 0.42 & \cca{60.61} & 0.36 & \cca{59.81} & 0.65 & \cca{60.39} & 0.38 & \cca{60.30} & 0.20 & 0.0097 & 0.0067 \\
 & & CoT & \textbf{\cca{68.49}} & 0.88 & \textbf{\cca{68.00}} & 0.44 & \textbf{\cca{67.69}} & 0.55 & \textbf{\cca{66.25}} & 0.47 & \textbf{\cca{66.91}} & 0.95 & \textbf{\cca{67.47}} & 0.29 & 0.0119 & 0.0030 \\
 \hline
\multirow[t]{4}{*}{MMLU} & \multirow[t]{4}{*}{gpt-3.5-turbo} & vanilla & \cca{59.40} & 0.62 & \cca{58.70} & 0.65 & \cca{59.33} & 0.69 & \cca{59.30} & 0.92 & \cca{60.03} & 0.76 & \cca{59.35} & 0.28 & 0.0030 & 0.0108 \\
 &  & CoT & \cca{57.68} & 0.75 & \cca{57.20} & 0.96 & \cca{58.53} & 0.63 & \cca{59.37} & 1.16 & \cca{58.72} & 0.67 & \cca{58.30} & 0.44 & -0.0068 & 0.0116 \\
 & \multirow[t]{4}{*}{gpt-4o-mini} & vanilla & \cca{67.05} & 0.64 & \cca{68.21} & 0.63 & \cca{68.09} & 0.65 & \cca{68.01} & 0.61 & \cca{68.85} & 0.44 & \cca{68.04} & 0.20 & -0.0052 & 0.0078 \\
 & & CoT & \textbf{\cca{79.56}} & 0.42 & \textbf{\cca{80.07}} & 0.50 & \textbf{\cca{80.93}} & 0.49 & \textbf{\cca{80.66}} & 0.55 & \textbf{\cca{81.21}} & 0.50 & \textbf{\cca{80.49}} & 0.31 & 0.0019 & 0.0144 \\
\bottomrule
\end{tabular}}
\caption{Precision evaluation in vanilla and CoT settings across two models and three datasets over 10 runs. We report the average and the standard deviation of the different runs for different metrics. Higher hit rates indicate greater precision, while lower hit rates suggest overprecision. Additionally, a high correlation (corr) between confidence levels and predicted interval lengths reflects stronger self-confidence awareness in the LLM. P.S. refers to prompting strategy. {  The results show a widespread overprecision across datasets and models. CoT prompting has mixed effects (i.e. it didn't improve GPT-3.5-Turbo), which contradicts previous studies on overestimation \cite{xiongcan}. The lack of significant change between the different levels of confidence in addition to lack of correlation between interval length and confidence level can be symptomatic of a) reduced understanding of internal confidence in LLMs b) inability to adjust self-confidence by following instructions.}}
\label{tab:overprec}
\end{table*}

\begin{table*}[!h]
    \centering
    \scalebox{0.8}{
    \setlength{\tabcolsep}{1.7pt}
        \begin{tabular}{llrrrrrrrrrrrrrr}
    \toprule
     &  & \multicolumn{2}{r}{hit-avg} & \multicolumn{2}{r}{hit@95\%} & \multicolumn{2}{r}{hit@90\%} & \multicolumn{2}{r}{hit@80\%} & \multicolumn{2}{r}{hit@70\%} & \multicolumn{2}{r}{hit@60\%} & \multicolumn{2}{r}{corr} \\
     &   & mean & std & mean & std & mean & std & mean & std & mean & std & mean & std & mean & std \\
    dataset & agg\_strategy &  &  &  &  &  &  &  &  &  &  &  &  &  &  \\
    \midrule
    \multirow[t]{4}{*}{FinQA} 
        & LWM & \cca{19.46} & 0.17 & \cca{22.58} & 0.41 & \cca{20.48} & 0.49 & \cca{19.36} & 0.35 & \cca{16.88} & 0.36 & \cca{17.99} & 0.42 & -0.0013 & 0.0028 \\
        & MIA & \cca{18.74} & 0.14 & \cca{21.84} & 0.34 & \cca{19.44} & 0.26 & \cca{18.87} & 0.36 & \cca{16.49} & 0.32 & \cca{17.09} & 0.39 & -0.0024 & 0.0022 \\
        & Union & \cca{33.88} & 0.16 & \cca{35.87} & 0.48 & \cca{34.54} & 0.38 & \cca{34.44} & 0.28 & \cca{31.89} & 0.33 & \cca{32.64} & 0.32 & 0.0013 & 0.0021 \\
        & iLWM & \cca{17.01} & 0.19 & \cca{19.29} & 0.28 & \cca{17.71} & 0.52 & \cca{17.05} & 0.38 & \cca{15.17} & 0.44 & \cca{15.83} & 0.36 & -0.0051 & 0.0018 \\
    \cline{1-16} \cline{2-16}
    \multirow[t]{4}{*}{Medical} 
        & LWM & \cca{56.03} & 0.18 & \cca{55.88} & 0.57 & \cca{55.48} & 0.47 & \cca{56.05} & 0.59 & \cca{55.99} & 0.42 & \cca{56.76} & 0.54 & 0.0113 & 0.0036 \\
        & MIA & \cca{56.53} & 0.18 & \cca{56.63} & 0.49 & \cca{56.64} & 0.53 & \cca{56.77} & 0.37 & \cca{56.14} & 0.23 & \cca{56.46} & 0.40 & 0.0133 & 0.0025 \\
        & Union & \cca{70.56} & 0.27 & \cca{71.09} & 0.31 & \cca{70.58} & 0.43 & \cca{70.66} & 0.53 & \cca{69.77} & 0.46 & \cca{70.69} & 0.30 & 0.0129 & 0.0036 \\
        & iLWM & \cca{51.12} & 0.14 & \cca{50.16} & 0.29 & \cca{50.36} & 0.53 & \cca{51.45} & 0.53 & \cca{51.12} & 0.45 & \cca{52.52} & 0.44 & 0.0127 & 0.0019 \\
    \cline{1-16} \cline{2-16}
    \multirow[t]{4}{*}{MMLU} 
        & LWM & \cca{58.39} & 0.13 & \cca{56.17} & 0.56 & \cca{55.59} & 0.58 & \cca{58.31} & 0.58 & \cca{59.87} & 0.51 & \cca{62.00} & 0.34 & -0.0047 & 0.0083 \\
        & MIA & \cca{65.20} & 0.25 & \cca{64.36} & 0.44 & \cca{64.23} & 0.45 & \cca{65.45} & 0.55 & \cca{65.50} & 0.68 & \cca{66.46} & 0.31 & -0.0032 & 0.0066 \\
        & Union & \cca{76.09} & 0.10 & \cca{75.82} & 0.31 & \cca{76.16} & 0.31 & \cca{75.87} & 0.32 & \cca{76.07} & 0.38 & \cca{76.54} & 0.29 & -0.0019 & 0.0054 \\
        & iLWM & \cca{46.74} & 0.23 & \cca{42.70} & 0.33 & \cca{42.80} & 0.63 & \cca{47.12} & 0.48 & \cca{48.86} & 0.51 & \cca{52.24} & 0.59 & 0.0007 & 0.0094 \\
    \bottomrule
\end{tabular}
    }
    \setlength{\tabcolsep}{1.7pt}
    \caption{Results of various aggregation-based refinement strategies on the GPT-4o-Mini model across different datasets in the single confidence setting, where sampling is performed separately for each confidence level. {  The results show that aggregation strategies generally don't improve overconfidence in LLMs in a single confidence setting except for the obvious Union strategy.}}
    \label{tab:refine_agg_res_single}
\end{table*}

\begin{table}[h]
    \centering
    \scalebox{0.8}{
    \setlength{\tabcolsep}{1.7pt}
    \begin{tabular}{lrrrrrrrrrr}
    \toprule
    agg\_strategy & \multicolumn{2}{r}{CWA} & \multicolumn{2}{r}{LWM} & \multicolumn{2}{r}{MIA} & \multicolumn{2}{r}{Union} & \multicolumn{2}{r}{iLWM} \\
     & mean & std & mean & std & mean & std & mean & std & mean & std \\
    dataset &  &  &  &  &  &  &  &  &  &  \\
    \midrule
    FinQA & \cca{19.58} & 0.37 & \cca{21.47} & 0.34 & \cca{19.23} & 0.28 & \cca{55.04} & 0.38 & \cca{16.02} & 0.27 \\
    Medical & \cca{52.84} & 0.48 & \cca{54.90} & 0.43 & \cca{53.18} & 0.42 & \cca{81.78} & 0.18 & \cca{44.88} & 0.39 \\
    MMLU & \cca{61.19} & 0.45 & \cca{62.31} & 0.43 & \cca{61.75} & 0.48 & \cca{84.62} & 0.28 & \cca{33.89} & 0.45 \\
    \bottomrule
    \end{tabular}}
    \caption{Performance of various aggregation-based refinement strategies on the GPT-4o-Mini model across different datasets in the mixed confidence setting, with sampling conducted separately for each confidence level. {  The results show that aggregation strategies generally don't improve overconfidence in LLMs in a mixed confidence setting except for the obvious Union strategy.}}
    \label{tab:refine_agg_res_mix}
\end{table}

\begin{table}[h]
    \centering
    \scalebox{0.8}{
    \setlength{\tabcolsep}{1.7pt}
\begin{tabular}{llrrrrrrr}
    \toprule
     &  & hit@95\% & hit@90\% & hit@80\% & hit@70\% & hit@60\% & hit-avg & corr \\
    dataset & kind &  &  &  &  &  &  &  \\
    \midrule
    \multirow[t]{2}{*}{FinQA} & chosen & \cca{20.56} & \cca{18.42} & \cca{17.73} & \cca{16.33} & \cca{17.36} & \cca{18.08} & -0.0170 \\
     & proposed & \cca{16.91} & \cca{15.75} & \cca{15.00} & \cca{13.26} & \cca{13.46} & \cca{14.88} & -0.0104 \\
    \cline{1-9}
    \multirow[t]{2}{*}{Medical} & chosen & \cca{59.52} & \cca{60.69} & \cca{61.06} & \cca{60.84} & \cca{61.08} & \cca{60.64} & 0.0191 \\
     & proposed & \cca{50.19} & \cca{52.43} & \cca{51.48} & \cca{50.73} & \cca{50.39} & \cca{51.04} & 0.0062 \\
    \cline{1-9}
    \multirow[t]{2}{*}{MMLU} & chosen & \cca{66.73} & \cca{66.92} & \cca{68.12} & \cca{67.08} & \cca{68.68} & \cca{67.51} & 0.0021 \\
     & proposed & \cca{59.75} & \cca{58.13} & \cca{59.78} & \cca{58.85} & \cca{57.78} & \cca{58.86} & 0.0030 \\
    \cline{1-9}
    \bottomrule
\end{tabular}
}
    \caption{\textbf{Self-refinement in the single confidence setting}: Self-refinement of answers generated using vanilla prompts from the GPT-4o-Mini model across different datasets, utilizing the GPT-4o-Mini LLM. For each question-answer pair, three possible answers are sampled from each confidence level. "Chosen" refers to the answers selected by the LLM from the proposed options, while "Proposed" represents the new interval suggested by the LLM. {  Self-refinement doesn't improve the performance in LLMs, which contradicts previous findings in cognitive science \cite{haran2010simple, moore2015wide} and LLMs applied to a mix of categorical and numerical data \cite{xiongcan}.}}
    \label{tab:selfrefine_single}
\end{table}
\begin{table}[h]
    \centering
    \scalebox{0.8}{
    \begin{tabular}{llr}
    \toprule
     &  & hit-avg \\
    dataset & kind &  \\
    \midrule
    \multirow[t]{2}{*}{FinQA} & chosen & \cca{18.54} \\
     & proposed & \cca{15.56} \\
    \cline{1-3}
    \multirow[t]{2}{*}{Medical} & chosen & \cca{60.59} \\
     & proposed & \cca{52.96} \\
    \cline{1-3}
    \multirow[t]{2}{*}{MMLU} & chosen & \cca{65.61} \\
     & proposed & \cca{59.13} \\
    \cline{1-3}
    \bottomrule
    \end{tabular}}
   \caption{\textbf{Self-refinement in the mixed confidence setting}: Using the GPT-4o-Mini model, self-refinement generates answers across datasets by sampling nine responses per question, regardless of confidence levels. "Chosen" refers to the LLM's selected answers, while "Proposed" represents the new intervals it suggests.}
    \label{tab:selfrefine_mix}
\end{table}
\begin{figure*}
    \centering
    \subfloat[FinQA|gpt-3.5-turbo]{\label{fig:finqa3}{\includegraphics[width=0.3\textwidth]{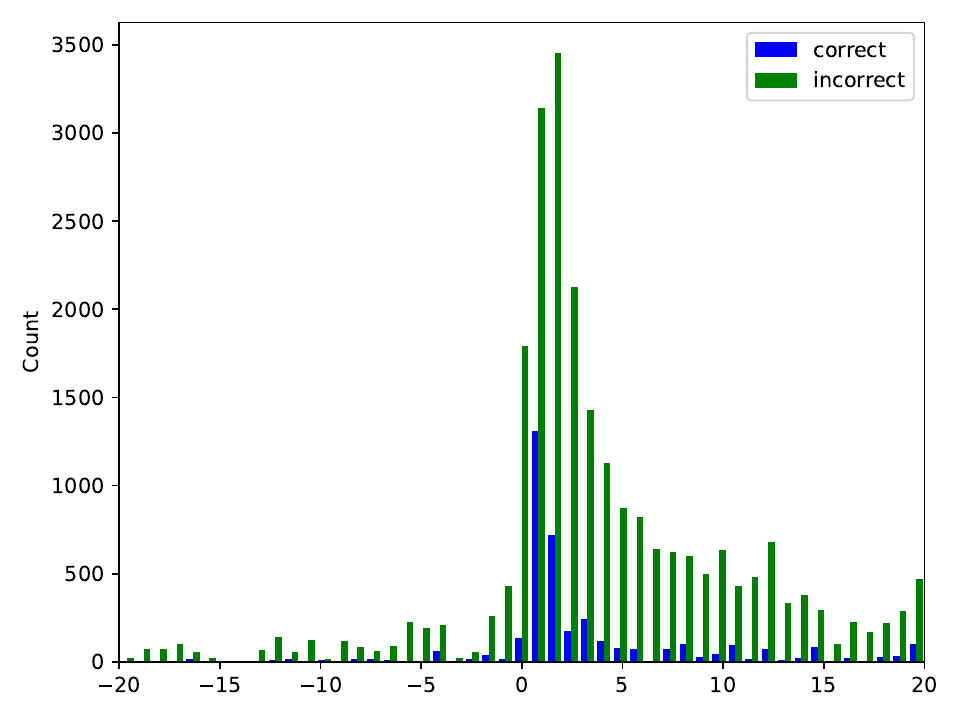}}}
    \hfill
    \subfloat[FinQA|gpt-4o-mini]{\label{fig:finqa4}{\includegraphics[width=0.3\textwidth]{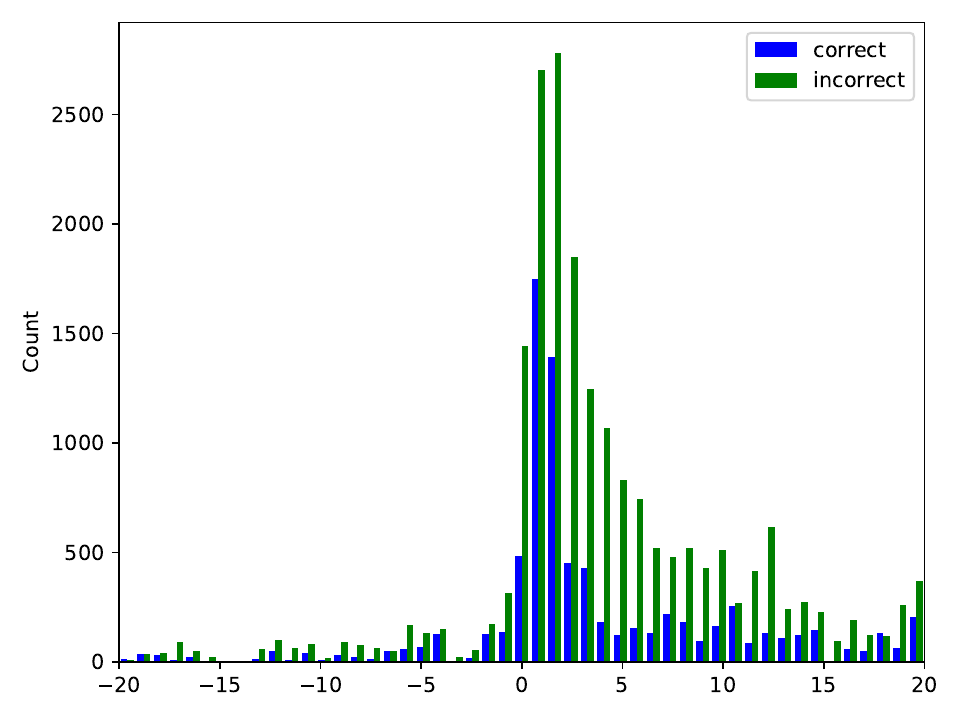}}}
    \hfill
    \subfloat[Medical|gpt-3.5-turbo]{\label{fig:medical3}{\includegraphics[width=0.3\textwidth]{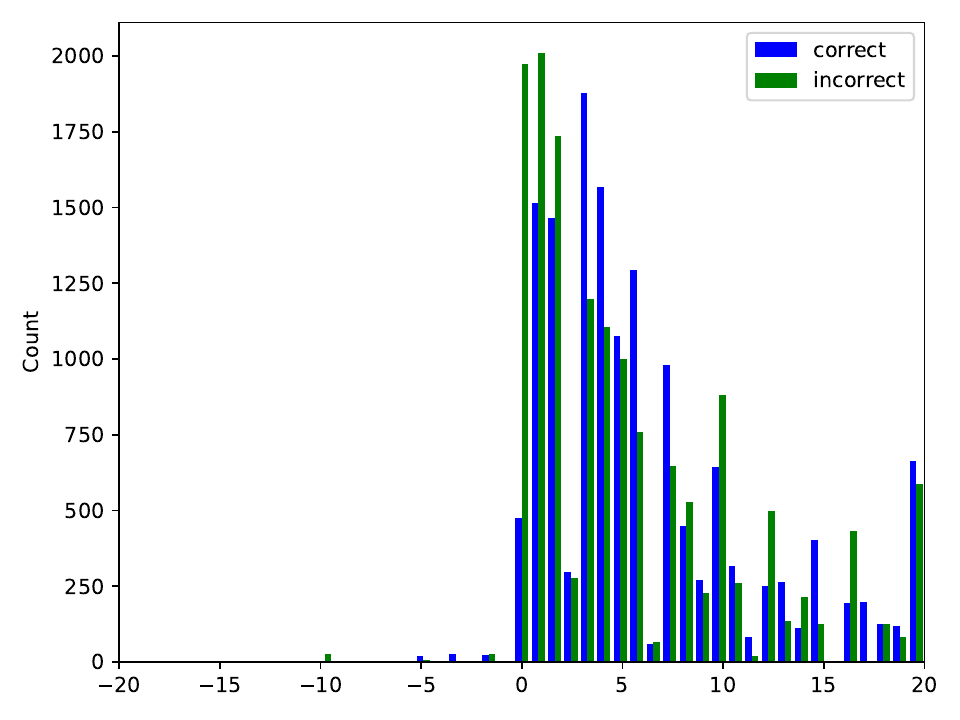}}}
    \hfill
    \subfloat[Medical|gpt-4o-mini]{\label{fig:medical4}{\includegraphics[width=0.3\textwidth]{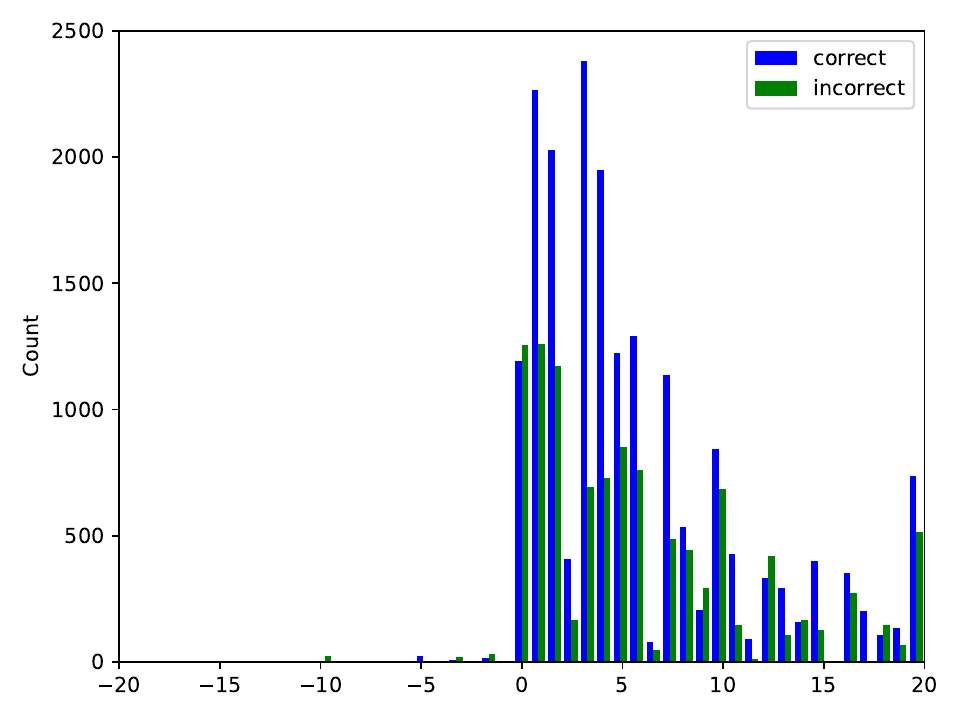}}}
    \hfill
    \subfloat[MMLU|gpt-3.5-turbo]{\label{fig:mmlu3}{\includegraphics[width=0.3\textwidth]{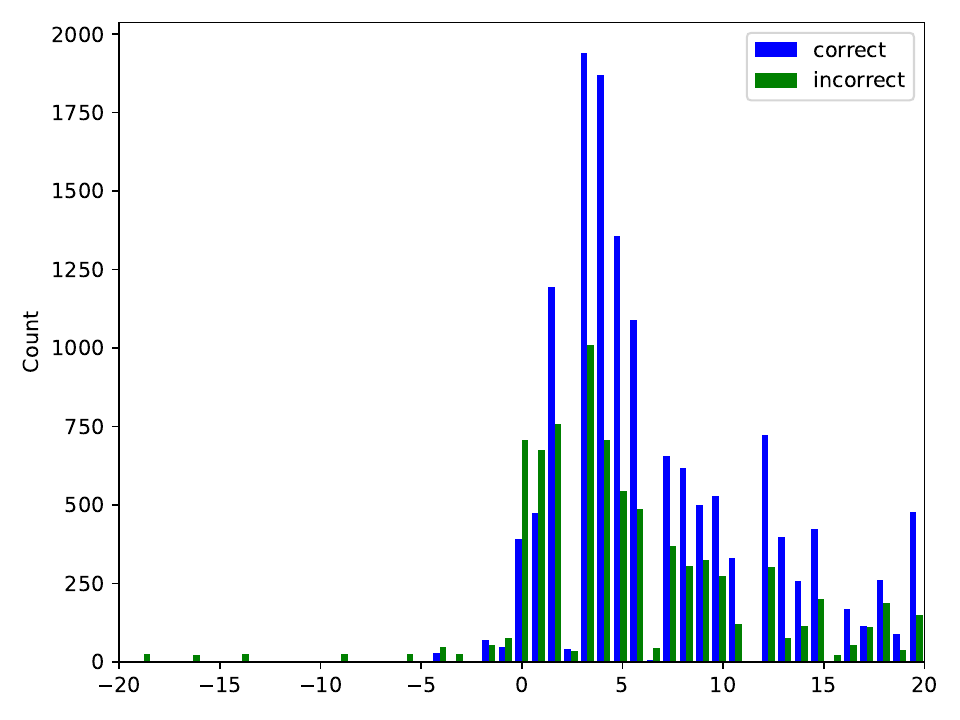}}}
    \hfill
    \subfloat[MMLU|gpt-4o-mini]{\label{fig:mmlu4}{\includegraphics[width=0.3\textwidth]{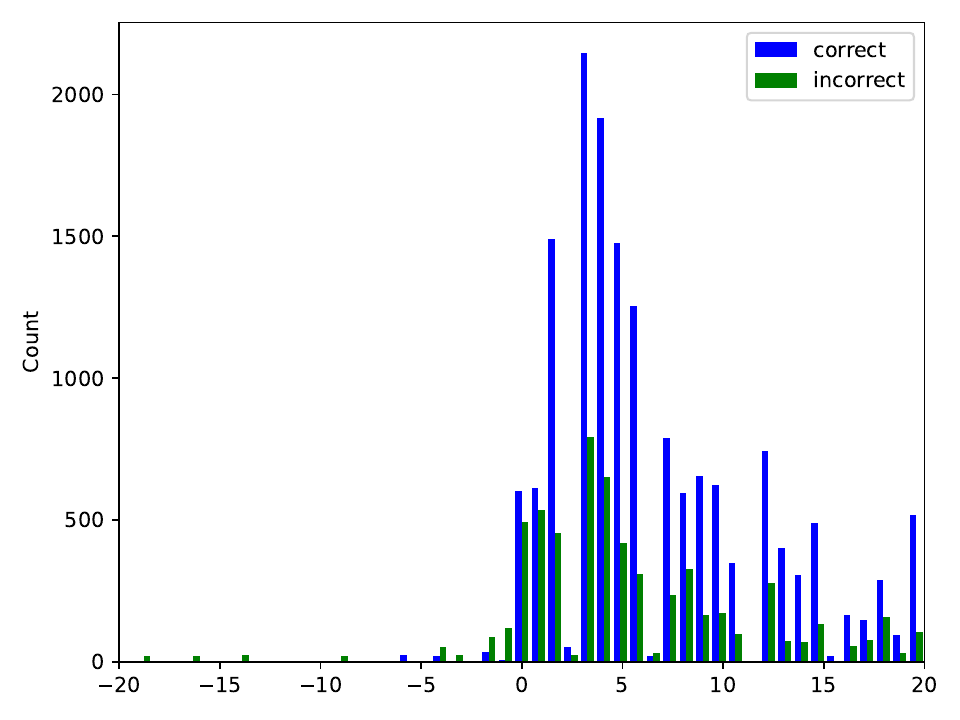}}}
    \hfill
    \caption{\textbf{Scale affect on precision}: These figures show the distribution of the hit average for different answers in the vanilla prompt setting for different models on different datasets. {  The figures demonstrate that the performance Is affected by the prompting strategy, the scale of the answer, and the task.}}
    \label{fig:scalevsprecision}
\end{figure*}

\subsection{LLMs are generally overprecise}
    Table \ref{tab:overprec} presents the results from Phase 1 (generation) for different models across various settings. All models exhibit overprecision to varying degrees of severity, as evidenced by the lack of calibration between the imposed confidence levels and the actual hit rates of the LLMs. CoT prompting significantly improves precision in the case of GPT-4o-Mini. However, CoT has a minimal impact on GPT-3.5-Turbo's performance and, in fact, slightly worsens its results for the MMLU dataset. These findings corroborate previous studies on overestimation in LLMs \cite{xiongcan, geng2024survey} and extend them to the overprecision aspect of overconfidence. Nonetheless, the lack of improvement with CoT prompts for GPT-3.5-Turbo contradicts the findings of \cite{xiongcan}, which observed positive effects of CoT prompts in the case of categorical data.
\subsection{LLMs' Confidence Does Not Correlate with Their Predictions}
    Table \ref{tab:overprec} demonstrates that the hit rate remains largely unchanged across different confidence levels for all models and datasets. Additionally, the lack of correlation between the lengths of the predicted intervals and the imposed confidence levels further supports {  this can be symptomatic of a) lack of understanding of the concept of confidence or b) inability to adjust self-confidence by following instructions. In appendix \ref{sec:dev_len_var}, this conclusion is substantiated by proposing two novel metrics to calculate the relative interval length. We found that the interval lengths effectively change depending on the level of knowledge that the LLM has. Consequently, this lack of correlation stems primarily from the inability of explored LLMs to control and regulate their internal states and self-confidence following instructions.}

\subsection{LLM Performance Is Affected by the Prompting Strategy, the Scale of the Answer, and the Task}
    Figure \ref{fig:scalevsprecision} demonstrates how the scale of ground truth answers influences LLM prediction accuracy. For example, in FinQA, predictions for answers near 0 tend to be more accurate, while accuracy declines for larger positive or negative values. Table \ref{tab:overprec} further emphasizes the impact of task type and prompting strategy on performance. Accuracy is significantly lower for specialized tasks such as FinQA and Medical, which require domain-specific knowledge, compared to general tasks like MMLU, which depend on broader knowledge without the need for specialized expertise.
\subsection{Refinement affects precision}
    \subsubsection{Aggregation}
    To validate the results, we performed 10 simulations, each involving random sampling of responses, and reported the average and standard deviation. In the single confidence setting, 3 trials per question-answer pair were sampled, whereas 9 trials were sampled in the mixed confidence setting. The results for the GPT-4o-Mini model using the vanilla prompt setting are presented in Tables \ref{tab:refine_agg_res_single} and \ref{tab:refine_agg_res_mix}.

    In the single confidence setting, the LWM, MIA, and Union aggregation strategies demonstrated improved performance compared to vanilla prompting, whereas the iLWM strategy resulted in reduced performance. For MMLU and Medical datasets, only the Union strategy showed significant improvement, primarily due to its reliance on larger intervals, which increases the likelihood of capturing the correct answer. Notably, the correlation between interval length and confidence level improved for the Medical dataset but showed no significant changes for MMLU or FinQA.

    In the mixed confidence setting (Table \ref{tab:refine_agg_res_mix}), the Union strategy consistently outperformed its single confidence counterpart, whereas the effects of other strategies on performance were mixed.
    
 \subsubsection{Self-refinement}
    Tables \ref{tab:selfrefine_single} and \ref{tab:selfrefine_mix} present the results of self-refinement in the single and mixed confidence settings, respectively. The LLM's choice of intervals did not improve the performance compared to the vanilla setting. Furthermore, the proposed intervals significantly reduced performance. This result contrasts sharply with findings in cognitive science, which show that when participants use other participants' responses to answer the same question, their performance improves \cite{haran2010simple, moore2015wide}. Additionally, this finding contradicts the results of \cite{xiongcan}, where the Self-Probing approach enhanced performance for mixed categorical and numerical data.
    
\section{Discussion}

    Throughout this study, several findings were established, either reinforcing previous research or challenging existing conclusions. We found that CoT reasoning improves the accuracy of certain models more effectively than others, emphasizing the varying adaptability of models to CoT-based prompts. Additionally, our results show that the numerical precision of LLMs is highly task-dependent, corroborating prior research in the context of mixed categorical and numerical data \cite{xiongcan}. However, our findings extend this understanding by demonstrating that precision is also influenced by the scale of the answer, indicating a more complex interaction between task characteristics and model outputs.
    
    Interestingly, our analysis of refinement strategies revealed inconsistent performance gains. In some cases, these strategies even degraded performance. This stands in stark contrast to prior work on mixed data for overestimation \cite{xiongcan, wen2024mitigating}, which reported consistent improvements through techniques like aggregation and self-probing. These discrepancies may arise from differences in dataset composition, task complexity, or implementation specifics, highlighting the need for further investigation into refinement strategies across a broader range of experimental conditions.
\section{Limitations and Future Work}
    1) \textit{Scope of Datasets}: This study primarily focused on two domains, finance and medicine, with some general knowledge tasks from MMLU. We believe this work can be further enhanced by extending experiments to other domains such as mathematics, law, biology, physics, and other fields involving numerical reasoning tasks. \\
    2) \textit{Scope of Models}: Due to budget constraints, we limited our experiments to two models. While these models exhibited varying behaviours, we aim to expand this study in the future by including a broader range of models to capture more diverse insights. \\
    3) \textit{Black-box setting}: The techniques proposed in this work are designed for black-box settings. However, we observed a lack of research on overprecision in white-box settings. Exploring this aspect could open new and interesting avenues for future research.
\section{Conclusion}
    This study addresses the underexplored phenomenon of overprecision in LLMs, providing key insights into their behaviour and limitations. Our findings demonstrate that LLMs are poorly calibrated for numerical tasks, with no observable correlation between interval length and confidence levels, indicating a lack of understanding of internal confidence. Numerical precision is shown to vary significantly depending on the task, the scale of the answer, and the prompting technique used. Refinement strategies, however, exhibit limited effectiveness, with self-refinement often resulting in decreased performance—contradicting prior findings in cognitive science and general LLM tasks. These results underscore the limitations of verbalized confidence elicitation and highlight the pressing need for more robust methods to study and mitigate overconfidence in LLMs.
\section*{Impact Statement}
    This work explores overprecision in large language models (LLMs), a robust aspect of overconfidence, contributing to a deeper understanding of their decision-making processes. The insights gained can inform the development of more calibrated and reliable AI systems, particularly in critical applications such as finance, medicine, and education, where overprecision could lead to significant societal or economic risks.

    From an ethical perspective, addressing overprecision in LLMs aligns with responsible AI development, reducing the potential harm caused by overconfident but incorrect predictions. For example, improving the calibration of LLMs can mitigate risks in areas like automated financial analysis or medical diagnostics, where erroneous confidence intervals could result in serious consequences.

    Future work expanding on this study could enhance transparency and accountability in AI systems by fostering more interpretable and dependable models. However, it is also crucial to acknowledge that improved confidence calibration could unintentionally enable misuse, such as deceptive practices or misinformation. To mitigate this, we encourage responsible deployment practices and further interdisciplinary research on the societal implications of AI systems.

    This work ultimately advances the field of machine learning by addressing the overlooked challenge of overprecision in LLMs, paving the way for more ethical and effective AI solutions.



\nocite{langley00}

\bibliography{main}
\bibliographystyle{icml2025}

\appendix
\section{Prompts}
    Table \ref{tab:inst_prompt_extra} shows a more complete version of the prompts in table \ref{tab:inst_prompt}.
\begin{table*}[]
       \centering
       \begin{tabular}{c|p{16em}|p{16em}}
            Instruction& Text & Objective\\
            \hline
            $\text{GEN}$&"please follow these instructions to ..."&General instructions that the LLM should follow\\
            $\text{CONF}_{c}$& "Please give us two numbers: a ‘lower bound’ and an ‘upper bound’. The ‘lower bound’ is a number so low that there is only a { $(100-\frac{c}{2})$\%} probability that the right answer is less than that.  Similarly, an ‘upper bound’ is a number so high that there is only a { $(100-\frac{c}{2})$\%} probability the right answer is more than that. In other words, you should be { $c$\%} sure that the answer falls between the lower and upper bounds" & Instructing the LLM on the level of confidence that it should have in its answer.\\
            $\text{CONFK}$&"The more unsure you are in your response the upper bound and the lower bound should be distant ..."&Giving the LLM general knowledge about confidence\\
            $\text{FORM}$&"your answer should have the following format ..."& Formating instructions that facilitate the parsing of the LLM output\\
            $\text{CoT}$& "give your step-by-step reasoning for why..."&Chain of Thought instructions that encourage the LLM to have better reasoning\\
            $\text{HINT}$&"I read in a book that the right answer is: { [lower\_bound, upper\_bound]}. Note that the hint is for reference only and may not be true."&Misleading hint given to the LLM to gouge its true confidence.\\
            $\text{QUES}(q_{i})$&"Question: { [$q_{i}$]}"&The question that the LLM should answer.\\
       \end{tabular}
       \caption{Sets of instructions that are used in the prompts. 'instruction' represents the abbreviation used in the paper for a particular set of instructions. 'Text' is the instruction text. 'objective' is the purpose of having that set of instructions.}
       \label{tab:inst_prompt_extra}
   \end{table*}
\section{Affects of Misleading Hints}
    Table \ref{tab:overprec_hints} presents the results of incorporating various hints into the prompts. The findings indicate that these hints can significantly enhance the performance of different models. This improvement can be attributed to the fact that the hints are generated around the expected answer, a technique adapted from \cite{xiongcan}. However, this approach may compromise the validity of the results in numerical settings, as it artificially boosts the accuracy of the LLM. A more effective strategy would involve ensuring that the proposed hints are as distant as possible from the correct answer, which would provide a more accurate assessment of overprecision.
    Additionally, the results show that different hints have varying impacts across datasets, further underscoring the importance of prompt optimization in mitigating overconfidence in LLMs.
    \begin{table*}[]
        \centering
         \scalebox{0.8}{
            \setlength{\tabcolsep}{1.7pt}
        \begin{tabular}{llllrrrrrrrrrrrrrr}
        \toprule
        &  &  &  & \multicolumn{2}{r}{hit-avg} & \multicolumn{2}{r}{hit@95\%} & \multicolumn{2}{r}{hit@90\%} & \multicolumn{2}{r}{hit@60\%} & \multicolumn{2}{r}{hit@70\%} & \multicolumn{2}{r}{hit@80\%} & \multicolumn{2}{r}{corr} \\
        &  &  &  & mean & std & mean & std & mean & std & mean & std & mean & std & mean & std & mean & std \\
        model & dataset & hint & P.S.&  &  &  &  &  &  &  &  &  &  &  &  &  &  \\
        \midrule
        \multirow[t]{18}{*}{gpt-3.5-turbo} & \multirow[t]{6}{*}{FinQA} & \multirow[t]{2}{*}{hint1} & vanilla & \cca{36.21} & 0.30 & \cca{37.62} & 0.70 & \cca{36.37} & 0.64 & \cca{35.65} & 0.69 & \cca{35.52} & 0.99 & \cca{35.90} & 0.70 & 0.0002 & 0.0056 \\
         &  &  & CoT & \cca{30.49} & 0.36 & \cca{28.91} & 0.73 & \cca{28.99} & 0.58 & \cca{33.60} & 0.54 & \cca{30.40} & 0.69 & \cca{30.55} & 0.64 & -0.0025 & 0.0073 \\
        \cline{3-18}
         &  & \multirow[t]{2}{*}{hint3} & vanilla & \cca{38.29} & 0.27 & \cca{40.97} & 0.55 & \cca{39.34} & 0.63 & \cca{35.72} & 0.83 & \cca{36.59} & 0.56 & \cca{38.80} & 0.71 & 0.0046 & 0.0043 \\
         &  &  & CoT & \cca{34.91} & 0.41 & \cca{35.51} & 0.69 & \cca{34.16} & 0.71 & \cca{35.83} & 0.80 & \cca{34.40} & 0.83 & \cca{34.67} & 0.53 & 0.0052 & 0.0079 \\
        \cline{3-18}
         &  & \multirow[t]{2}{*}{hint8} & vanilla & \cca{36.98} & 0.23 & \cca{38.71} & 0.55 & \cca{36.77} & 0.54 & \cca{35.37} & 0.60 & \cca{36.85} & 0.76 & \cca{37.18} & 0.72 & 0.0028 & 0.0066 \\
         &  &  & CoT & \cca{37.75} & 0.26 & \cca{37.80} & 0.85 & \cca{35.22} & 0.29 & \cca{39.45} & 0.45 & \cca{37.76} & 0.56 & \cca{38.54} & 0.71 & 0.0090 & 0.0075 \\
        \cline{2-18} \cline{3-18}
         & \multirow[t]{6}{*}{Medical} & \multirow[t]{2}{*}{hint1} & vanilla & \cca{58.07} & 0.27 & \cca{58.59} & 0.60 & \cca{57.88} & 0.72 & \cca{57.47} & 0.91 & \cca{57.41} & 0.64 & \cca{59.02} & 0.63 & 0.0088 & 0.0045 \\
         &  &  & CoT & \cca{59.53} & 0.40 & \cca{59.89} & 0.43 & \cca{59.85} & 0.81 & \cca{58.55} & 1.08 & \cca{59.47} & 0.59 & \cca{59.90} & 0.68 & 0.0072 & 0.0091 \\
        \cline{3-18}
         &  & \multirow[t]{2}{*}{hint3} & vanilla & \cca{57.61} & 0.23 & \cca{59.35} & 0.73 & \cca{58.84} & 0.93 & \cca{55.52} & 0.95 & \cca{55.93} & 0.74 & \cca{58.40} & 0.96 & 0.0063 & 0.0063 \\
         &  &  & CoT & \cca{58.59} & 0.43 & \cca{59.78} & 1.24 & \cca{58.58} & 0.98 & \cca{57.57} & 0.58 & \cca{58.64} & 0.79 & \cca{58.40} & 1.20 & -0.0015 & 0.0092 \\
        \cline{3-18}
         &  & \multirow[t]{2}{*}{hint8} & vanilla & \cca{56.90} & 0.38 & \cca{57.67} & 0.90 & \cca{58.61} & 0.68 & \cca{56.69} & 0.84 & \cca{55.77} & 1.19 & \cca{55.75} & 0.76 & 0.0020 & 0.0040 \\
         &  &  & CoT & \cca{59.26} & 0.24 & \cca{60.76} & 0.74 & \cca{59.48} & 1.06 & \cca{57.80} & 0.74 & \cca{59.68} & 0.84 & \cca{58.58} & 0.58 & 0.0035 & 0.0076 \\
        \cline{2-18} \cline{3-18}
         & \multirow[t]{6}{*}{MMLU} & \multirow[t]{2}{*}{hint1} & vanilla & \cca{58.12} & 0.45 & \cca{58.80} & 0.77 & \cca{57.88} & 0.76 & \cca{57.64} & 0.89 & \cca{58.07} & 1.31 & \cca{58.23} & 0.88 & -0.0030 & 0.0129 \\
         &  &  & CoT & \cca{59.08} & 0.51 & \cca{58.41} & 1.12 & \cca{58.31} & 0.89 & \cca{59.71} & 1.09 & \cca{59.85} & 0.83 & \cca{59.10} & 1.19 & -0.0014 & 0.0138 \\
        \cline{3-18}
         &  & \multirow[t]{2}{*}{hint3} & vanilla & \cca{57.50} & 0.49 & \cca{58.02} & 0.87 & \cca{58.11} & 0.91 & \cca{56.23} & 1.10 & \cca{57.52} & 0.64 & \cca{57.63} & 1.03 & -0.0048 & 0.0100 \\
         &  &  & CoT & \cca{58.75} & 0.56 & \cca{59.09} & 0.73 & \cca{57.90} & 0.79 & \cca{59.63} & 0.86 & \cca{58.99} & 1.32 & \cca{58.13} & 1.04 & -0.0040 & 0.0127 \\
        \cline{3-18}
         &  & \multirow[t]{2}{*}{hint8} & vanilla & \cca{57.50} & 0.63 & \cca{59.20} & 1.05 & \cca{57.76} & 0.95 & \cca{55.80} & 0.83 & \cca{57.27} & 0.90 & \cca{57.50} & 0.95 & 0.0006 & 0.0060 \\
         &  &  & CoT & \cca{59.17} & 0.38 & \cca{59.03} & 1.05 & \cca{57.76} & 0.85 & \cca{59.42} & 0.77 & \cca{60.08} & 0.86 & \cca{59.54} & 0.63 & -0.0067 & 0.0117 \\
        \cline{1-18} \cline{2-18} \cline{3-18}
        \multirow[t]{18}{*}{gpt-4o-mini} & \multirow[t]{6}{*}{FinQA} & \multirow[t]{2}{*}{hint1} & vanilla & \cca{49.64} & 0.41 & \cca{50.71} & 0.62 & \cca{49.59} & 0.98 & \cca{50.00} & 0.85 & \cca{48.24} & 0.63 & \cca{49.67} & 0.52 & -0.0010 & 0.0027 \\
         &  &  & CoT & \cca{45.90} & 0.42 & \cca{46.81} & 0.79 & \cca{46.72} & 0.68 & \cca{44.81} & 0.93 & \cca{44.58} & 0.50 & \cca{46.58} & 0.62 & 0.0030 & 0.0071 \\
        \cline{3-18}
         &  & \multirow[t]{2}{*}{hint3} & vanilla & \cca{44.56} & 0.30 & \cca{44.20} & 0.65 & \cca{43.99} & 0.59 & \cca{45.52} & 0.65 & \cca{44.27} & 0.77 & \cca{44.84} & 0.76 & -0.0005 & 0.0031 \\
         &  &  & CoT & \cca{44.65} & 0.28 & \cca{46.00} & 0.58 & \cca{44.16} & 0.62 & \cca{44.16} & 0.77 & \cca{44.61} & 0.57 & \cca{44.29} & 0.84 & -0.0024 & 0.0030 \\
        \cline{3-18}
         &  & \multirow[t]{2}{*}{hint8} & vanilla & \cca{47.16} & 0.38 & \cca{47.22} & 0.79 & \cca{47.83} & 0.65 & \cca{47.16} & 0.56 & \cca{46.66} & 0.75 & \cca{46.91} & 0.52 & -0.0005 & 0.0045 \\
         &  &  & CoT & \cca{45.37} & 0.32 & \cca{46.05} & 0.65 & \cca{45.35} & 0.47 & \cca{44.69} & 0.87 & \cca{45.31} & 0.81 & \cca{45.46} & 0.68 & -0.0081 & 0.0076 \\
        \cline{2-18} \cline{3-18}
         & \multirow[t]{6}{*}{Medical} & \multirow[t]{2}{*}{hint1} & vanilla & \cca{69.20} & 0.46 & \cca{69.76} & 0.66 & \cca{69.59} & 0.80 & \cca{69.65} & 0.67 & \cca{68.05} & 0.71 & \cca{68.94} & 0.85 & 0.0006 & 0.0061 \\
         &  &  & CoT & \cca{74.17} & 0.21 & \cca{75.04} & 0.73 & \cca{75.07} & 0.65 & \cca{73.84} & 0.52 & \cca{72.94} & 0.44 & \cca{73.97} & 0.77 & -0.0022 & 0.0091 \\
        \cline{3-18}
         &  & \multirow[t]{2}{*}{hint3} & vanilla & \cca{60.52} & 0.50 & \cca{60.40} & 0.85 & \cca{60.72} & 1.04 & \cca{61.52} & 0.83 & \cca{58.98} & 0.78 & \cca{60.99} & 1.16 & 0.0015 & 0.0053 \\
         &  &  & CoT & \cca{69.94} & 0.33 & \cca{70.03} & 0.69 & \cca{70.59} & 0.82 & \cca{68.79} & 0.86 & \cca{69.66} & 0.77 & \cca{70.63} & 0.58 & -0.0023 & 0.0092 \\
        \cline{3-18}
         &  & \multirow[t]{2}{*}{hint8} & vanilla & \cca{66.29} & 0.29 & \cca{67.47} & 0.84 & \cca{67.09} & 0.91 & \cca{66.49} & 1.13 & \cca{63.54} & 0.81 & \cca{66.84} & 0.92 & -0.0023 & 0.0045 \\
         &  &  & CoT & \cca{72.34} & 0.33 & \cca{73.06} & 0.64 & \cca{72.05} & 0.50 & \cca{72.46} & 0.70 & \cca{71.92} & 0.51 & \cca{72.21} & 0.68 & -0.0024 & 0.0121 \\
        \cline{2-18} \cline{3-18}
         & \multirow[t]{6}{*}{MMLU} & \multirow[t]{2}{*}{hint1} & vanilla & \cca{68.81} & 0.39 & \cca{68.04} & 1.16 & \cca{68.82} & 0.70 & \cca{69.75} & 0.96 & \cca{68.24} & 0.67 & \cca{69.21} & 0.67 & 0.0008 & 0.0048 \\
         &  &  & CoT & \cca{81.91} & 0.32 & \cca{82.16} & 0.73 & \cca{81.71} & 0.65 & \cca{82.32} & 0.61 & \cca{81.17} & 0.91 & \cca{82.19} & 0.65 & 0.0073 & 0.0034 \\
        \cline{3-18}
         &  & \multirow[t]{2}{*}{hint3} & vanilla & \cca{63.68} & 0.55 & \cca{63.23} & 1.06 & \cca{63.57} & 1.02 & \cca{64.91} & 0.97 & \cca{62.68} & 1.33 & \cca{64.02} & 0.83 & 0.0016 & 0.0049 \\
         &  &  & CoT & \cca{76.85} & 0.41 & \cca{76.67} & 1.06 & \cca{76.20} & 0.70 & \cca{77.65} & 0.90 & \cca{76.46} & 0.52 & \cca{77.25} & 0.83 & -0.0091 & 0.0042 \\
        \cline{3-18}
         &  & \multirow[t]{2}{*}{hint8} & vanilla & \cca{65.60} & 0.24 & \cca{65.34} & 1.02 & \cca{65.83} & 0.81 & \cca{65.34} & 0.98 & \cca{64.91} & 0.98 & \cca{66.56} & 1.11 & -0.0037 & 0.0050 \\
         &  &  & CoT & \cca{78.68} & 0.24 & \cca{78.82} & 1.20 & \cca{78.49} & 0.61 & \cca{78.14} & 0.60 & \cca{79.35} & 0.62 & \cca{78.60} & 0.73 & 0.0057 & 0.0078 \\
        \cline{1-18} \cline{2-18} \cline{3-18}
        \bottomrule
        \end{tabular}}
        \caption{Results of using different hints. hint1 signifies the following instruction "Hint: I think the answer should be ...". hint3 signifies the following instruction: "Hint: The textbook shows that answer is ...". hint8 signifies the following instruction: "Hint: I vaguely remember the answer is ...". These hints are followed by a randomly generated interval where the answer may or may not fall. P.S. stands for prompting strategy.}
        \label{tab:overprec_hints}
    \end{table*}
\section{Effects of number of possible answers on self-refinement}
    
\begin{figure*}
    \centering
    \subfloat[Single|FinQA|Chosen]{\label{fig:finqa3}{\includegraphics[width=0.25\textwidth]{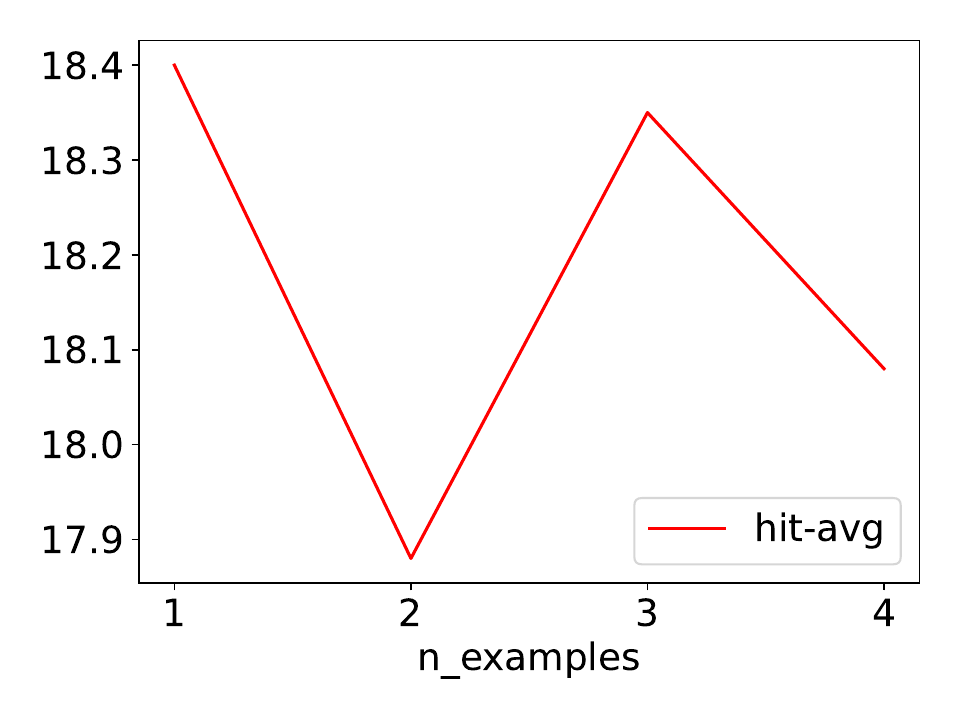}}}
    \hfill
    \subfloat[Single|FinQA|Proposed]{\label{fig:finqa4}{\includegraphics[width=0.25\textwidth]{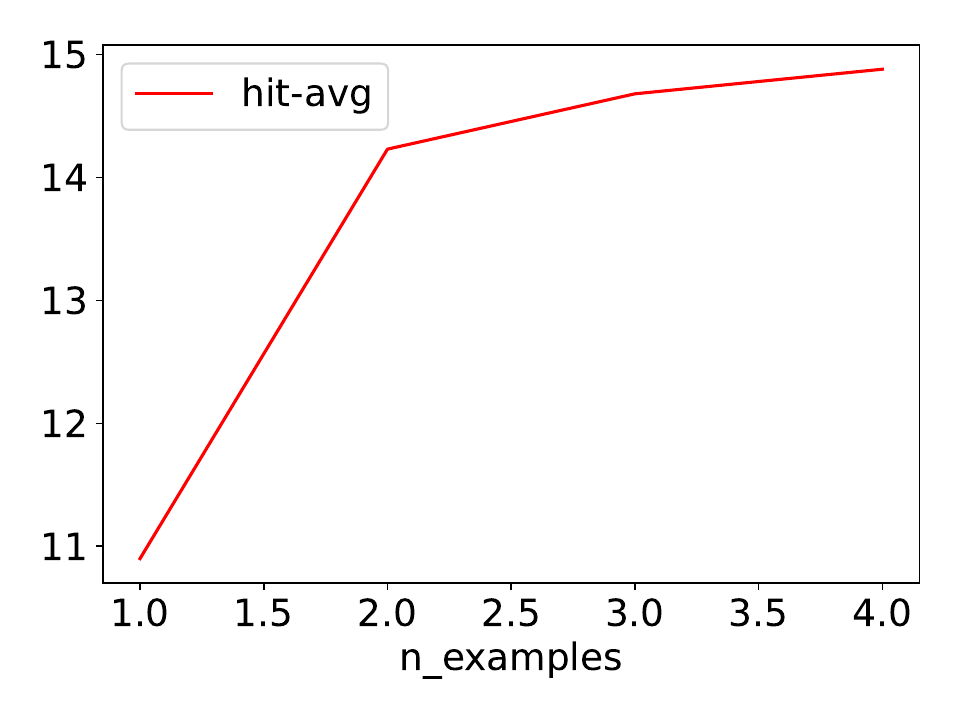}}}
    \hfill
    \subfloat[Single|Medical|Chosen]{\label{fig:finqa3}{\includegraphics[width=0.25\textwidth]{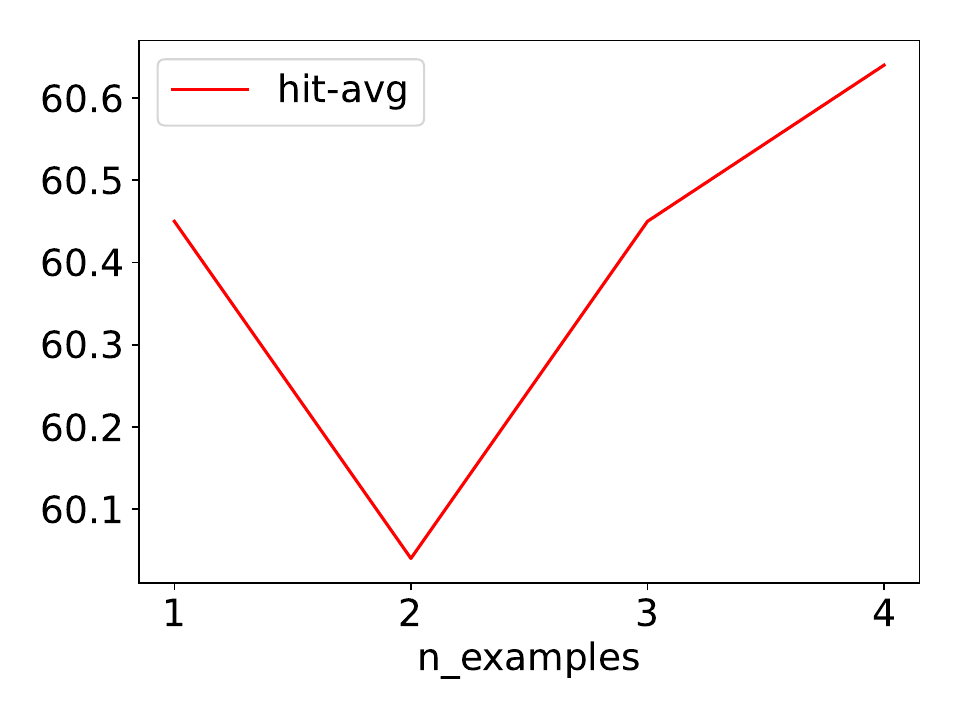}}}
    \hfill
    \subfloat[Single|Medical|Proposed]{\label{fig:finqa4}{\includegraphics[width=0.25\textwidth]{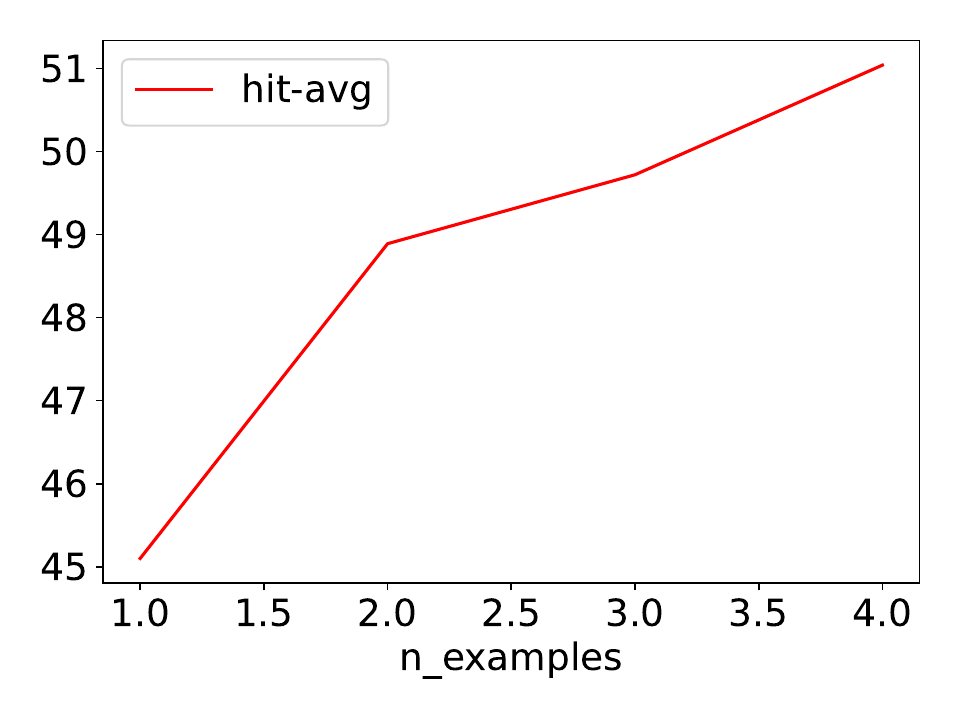}}}
    \hfill
    \subfloat[Single|MMLU|Chosen]{\label{fig:finqa3}{\includegraphics[width=0.25\textwidth]{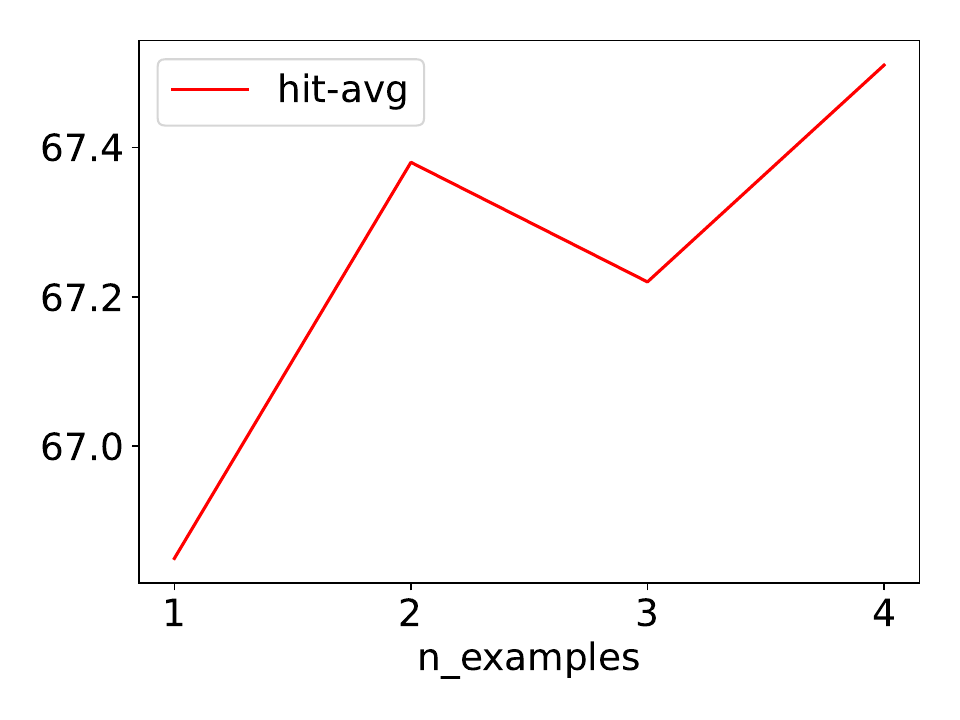}}}
    \hfill
    \subfloat[Single|MMLU|Proposed]{\label{fig:finqa4}{\includegraphics[width=0.25\textwidth]{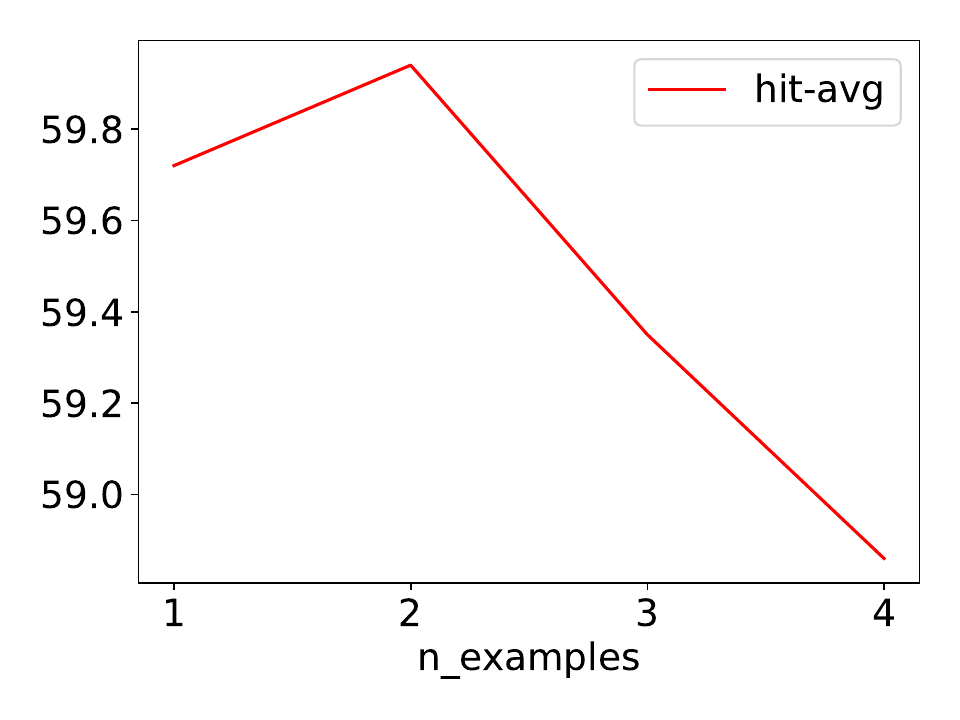}}}
    \hfill
    \subfloat[Mixed|FinQA|Chosen]{\label{fig:finqa3}{\includegraphics[width=0.25\textwidth]{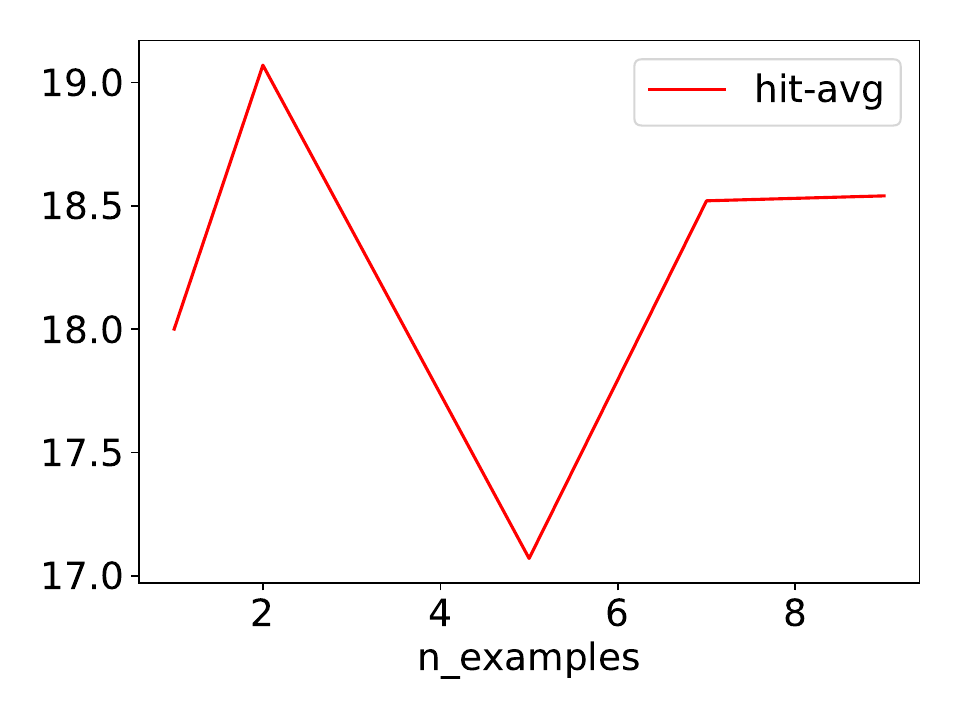}}}
    \hfill
    \subfloat[Mixed|FinQA|Proposed]{\label{fig:finqa4}{\includegraphics[width=0.25\textwidth]{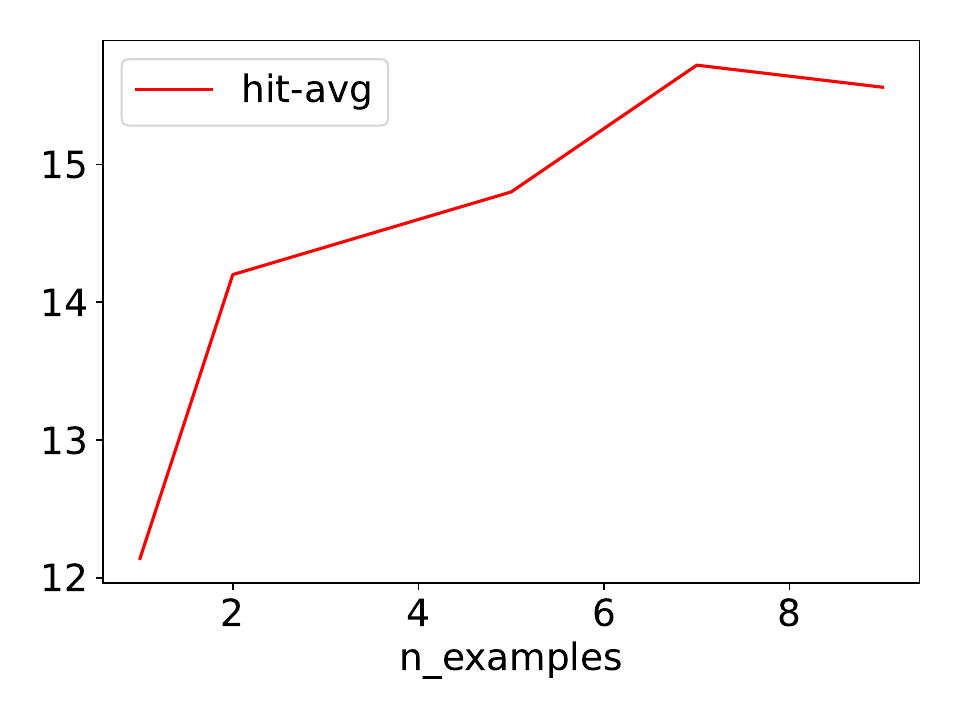}}}
    \hfill
    \subfloat[Mixed|Medical|Chosen]{\label{fig:finqa3}{\includegraphics[width=0.25\textwidth]{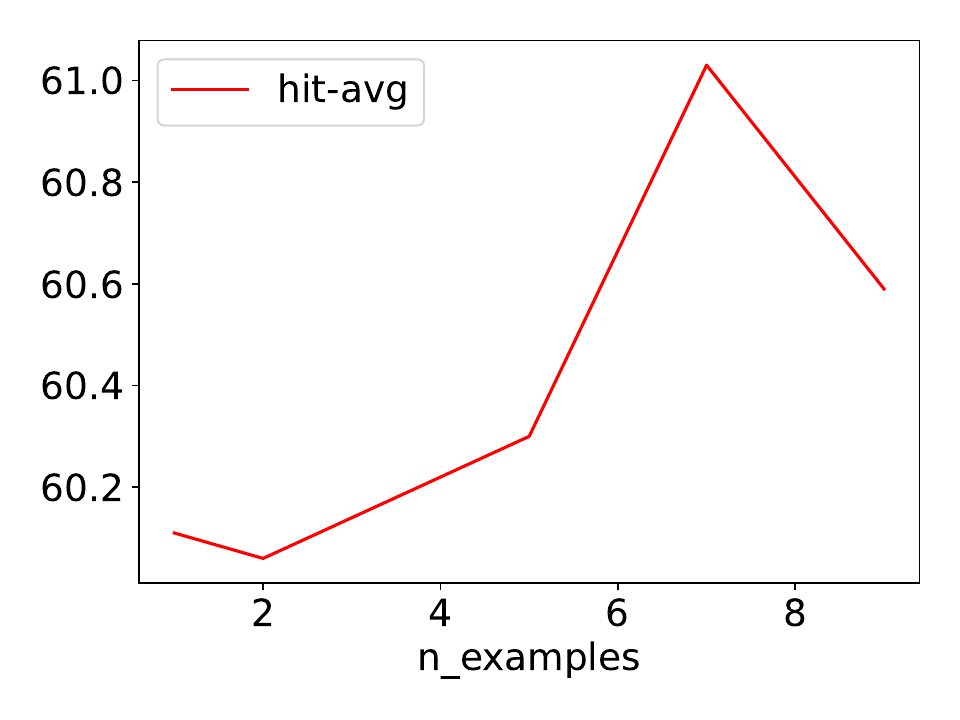}}}
    \hfill
    \subfloat[Mixed|Medical|Proposed]{\label{fig:finqa4}{\includegraphics[width=0.25\textwidth]{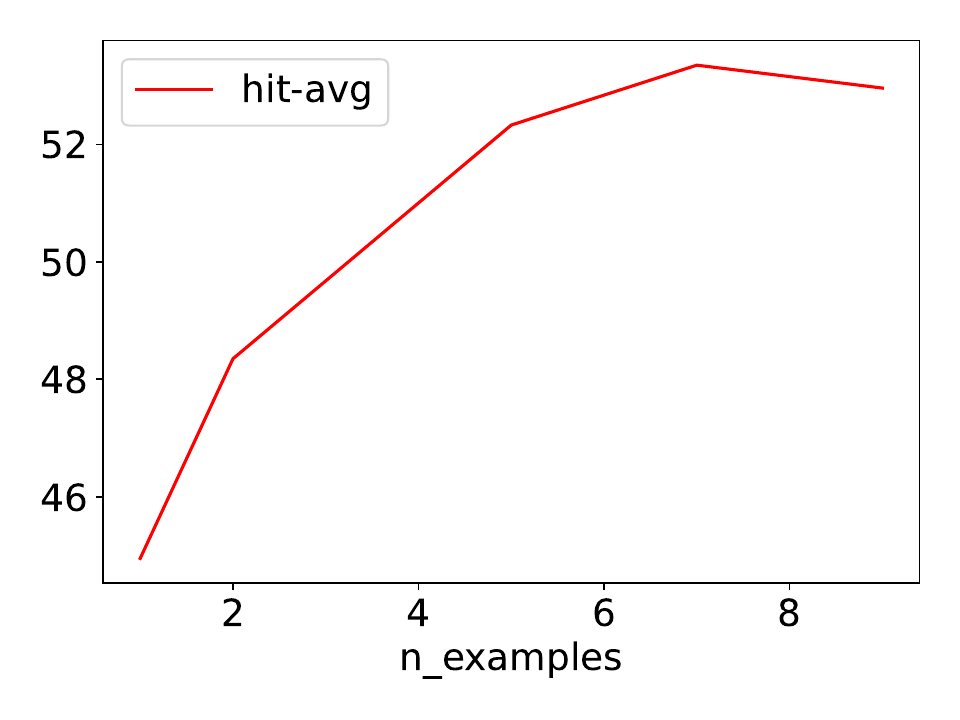}}}
    \hfill
    \subfloat[Mixed|MMLU|Chosen]{\label{fig:finqa3}{\includegraphics[width=0.25\textwidth]{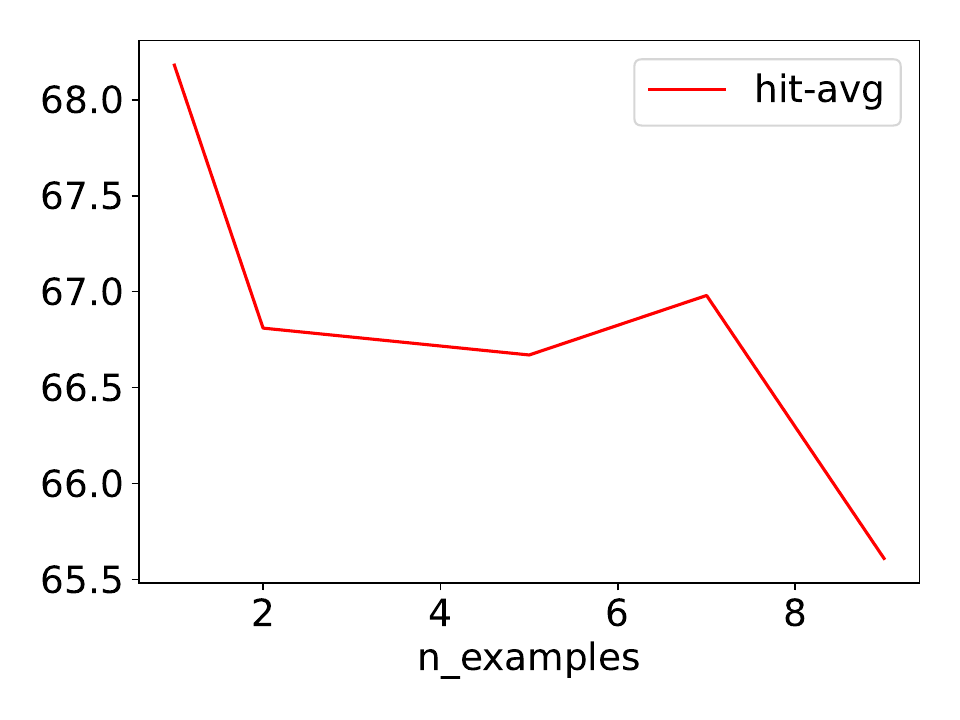}}}
    \hfill
    \subfloat[Mixed|MMLU|Proposed]{\label{fig:finqa4}{\includegraphics[width=0.25\textwidth]{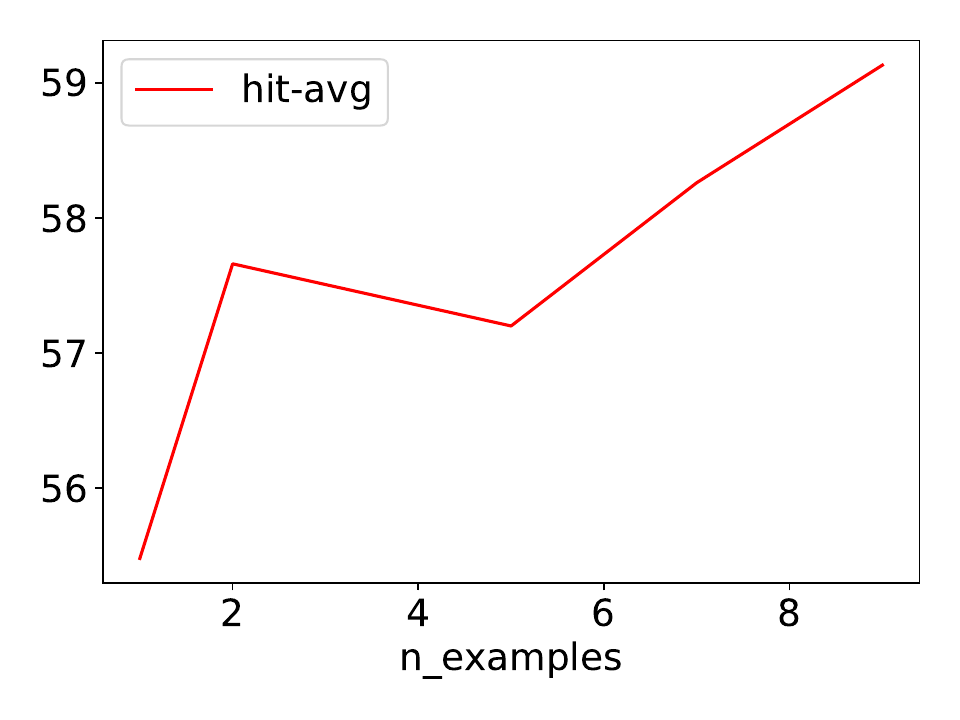}}}
    \caption{The hit average metric as a function of the number of examples provided in the self-refinement prompt. The titles of the subfigures are organized as follows: [setting][dataset][kind]. The setting can either be Single or mixed (refer to the experimental protocol for more detail). The kind can either be "chosen" for answers that were selected by the LLM to be the most correct. The kind can also be "proposed" for the answers that were proposed by the LLM but didn't exist in the provided examples.}
    \label{fig:selfrevevol}
\end{figure*}

\begin{figure*}
    \centering
    \subfloat[FinQA/GPT-3.5/Vanilla]{\label{fig:finqa3}{\includegraphics[width=0.25\textwidth]{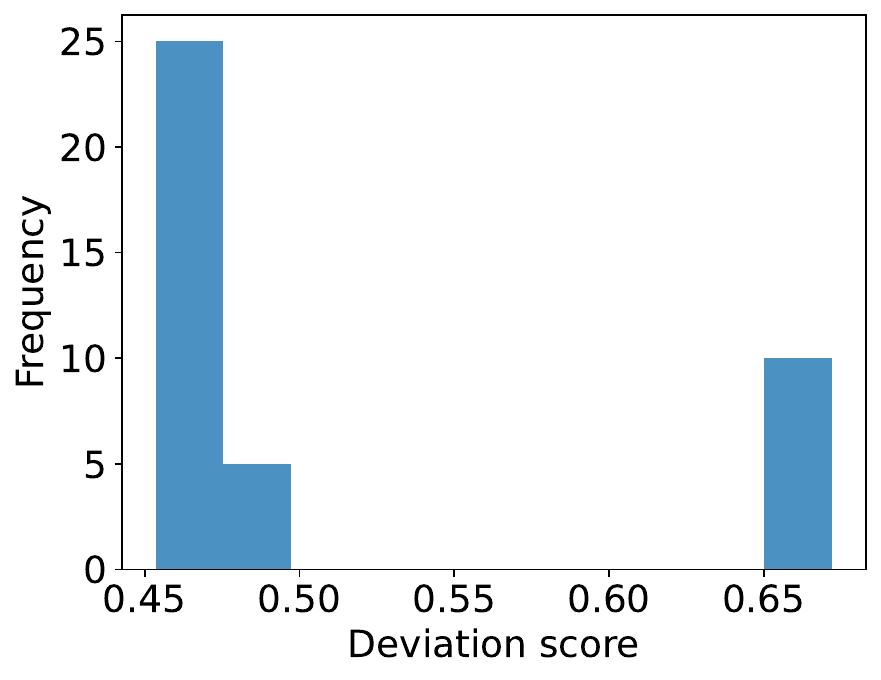}}}
    \hfill
    \subfloat[FinQA/GPT-3.5/CoT]{\label{fig:finqa3}{\includegraphics[width=0.25\textwidth]{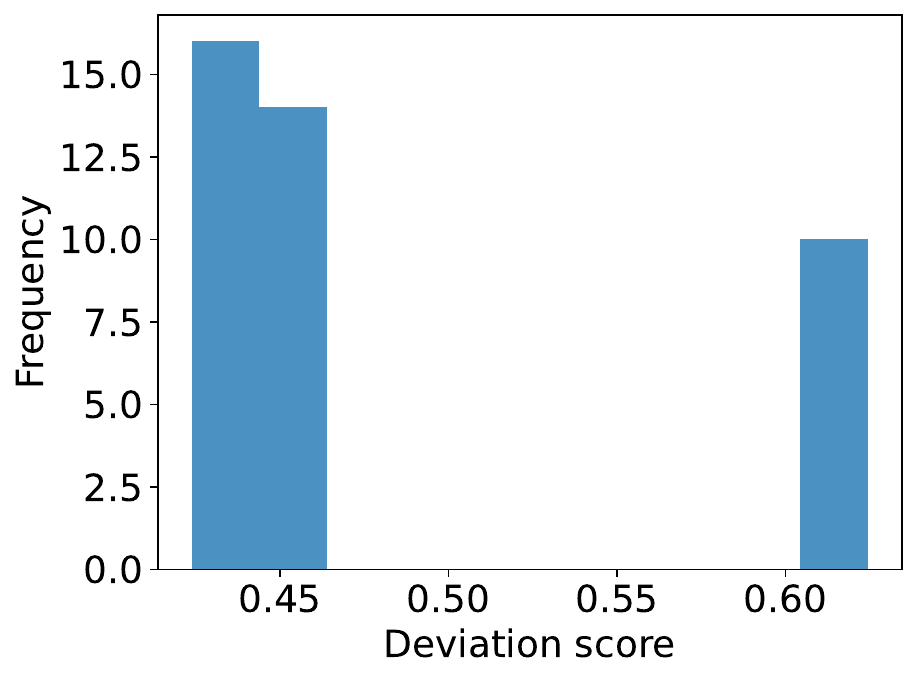}}}
    \hfill
    \subfloat[FinQA/GPT-4o/Vanilla]{\label{fig:finqa3}{\includegraphics[width=0.25\textwidth]{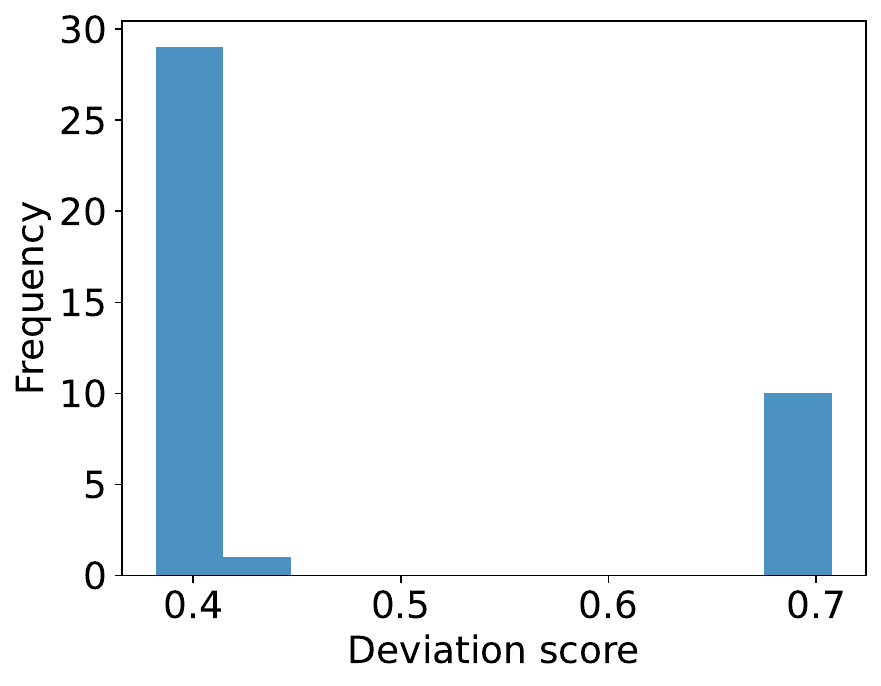}}}
    \hfill
    \subfloat[FinQA/GPT-4o/CoT]{\label{fig:finqa3}{\includegraphics[width=0.25\textwidth]{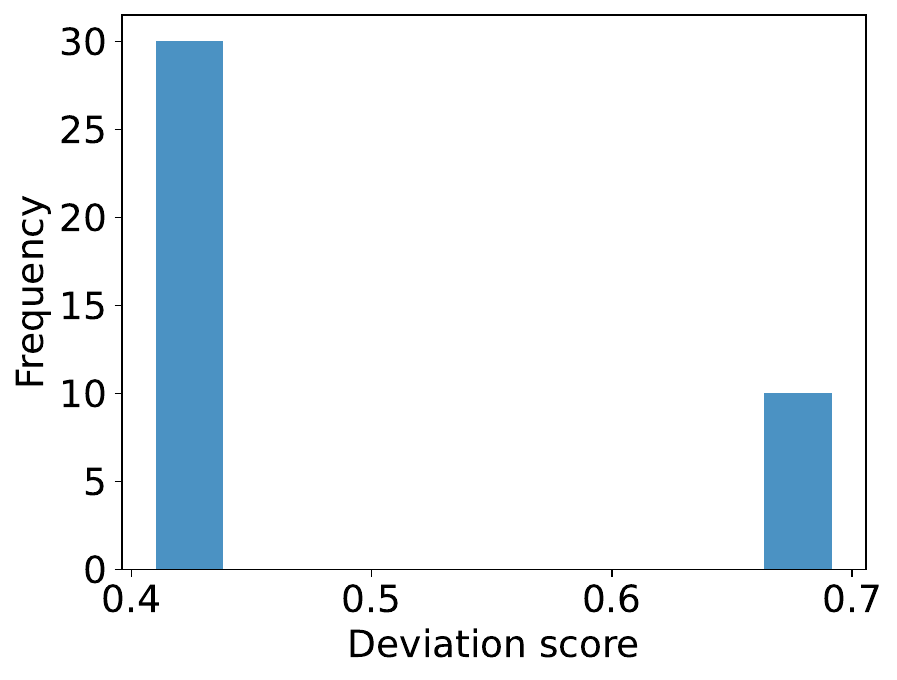}}}
    \hfill
    \subfloat[Medical/GPT-3.5/Vanilla]{\label{fig:finqa3}{\includegraphics[width=0.25\textwidth]{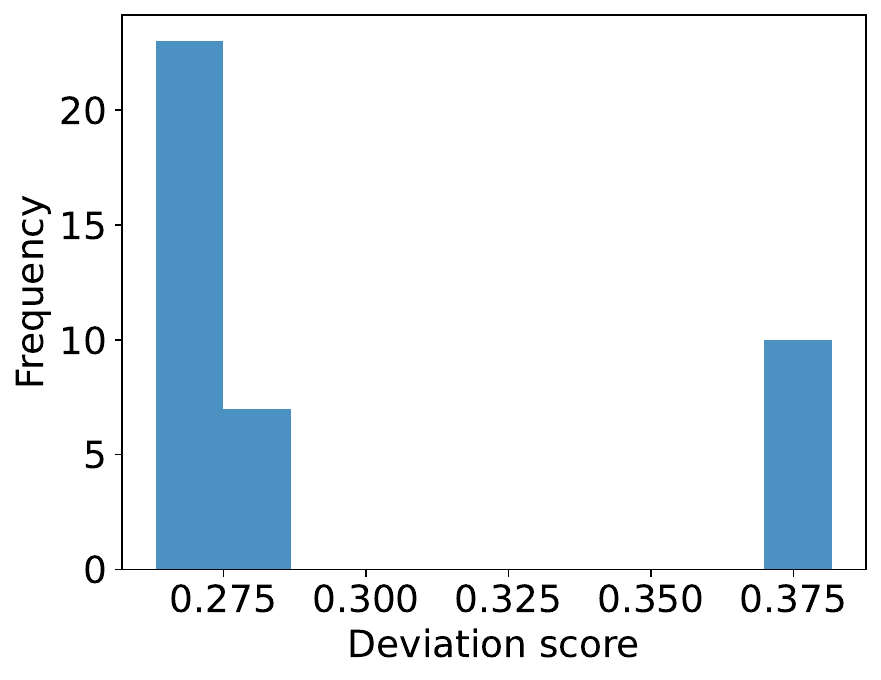}}}
    \hfill
    \subfloat[Medical/GPT-3.5/CoT]{\label{fig:finqa3}{\includegraphics[width=0.25\textwidth]{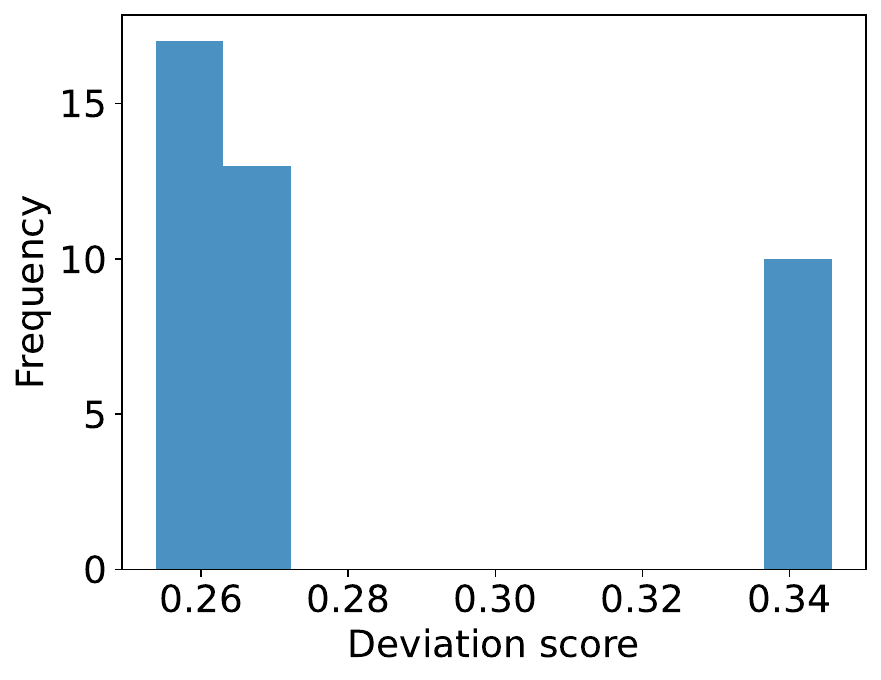}}}
    \hfill
    \subfloat[Medical/GPT-4o/Vanilla]{\label{fig:finqa3}{\includegraphics[width=0.25\textwidth]{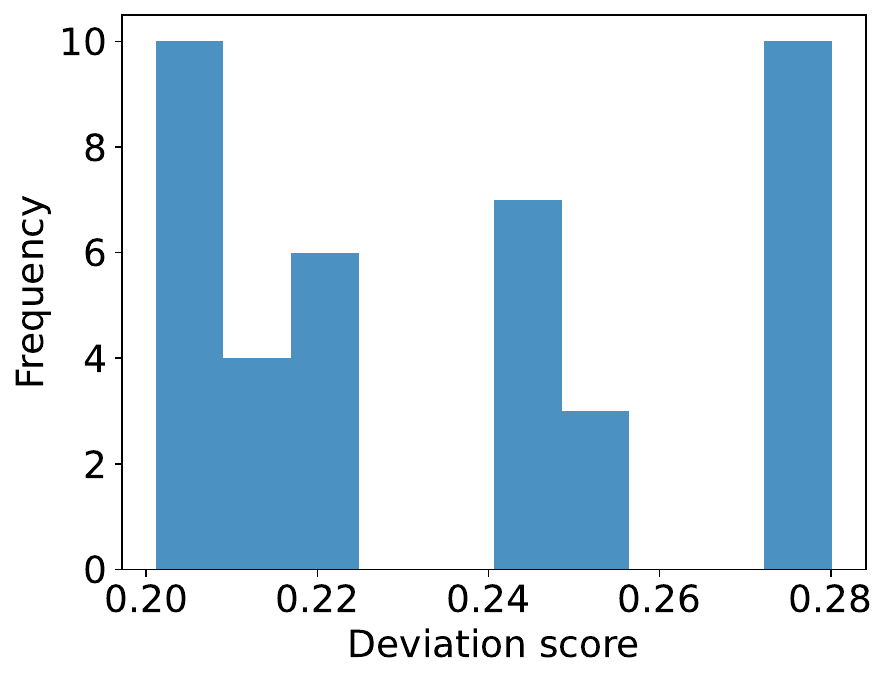}}}
    \hfill
    \subfloat[Medical/GPT-4o/CoT]{\label{fig:finqa3}{\includegraphics[width=0.25\textwidth]{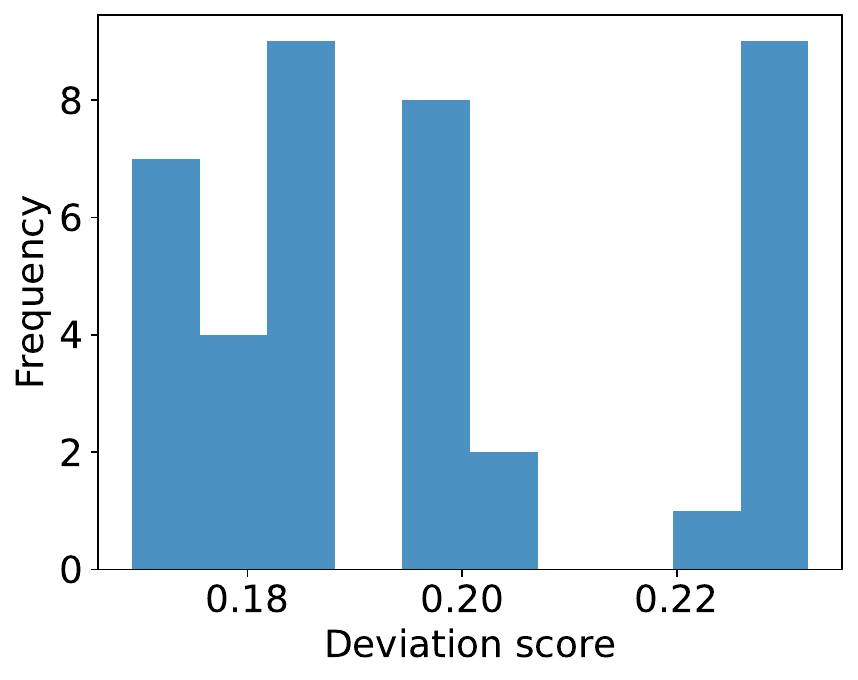}}}
    \hfill
    \subfloat[MMLU/GPT-3.5/Vanilla]{\label{fig:finqa3}{\includegraphics[width=0.25\textwidth]{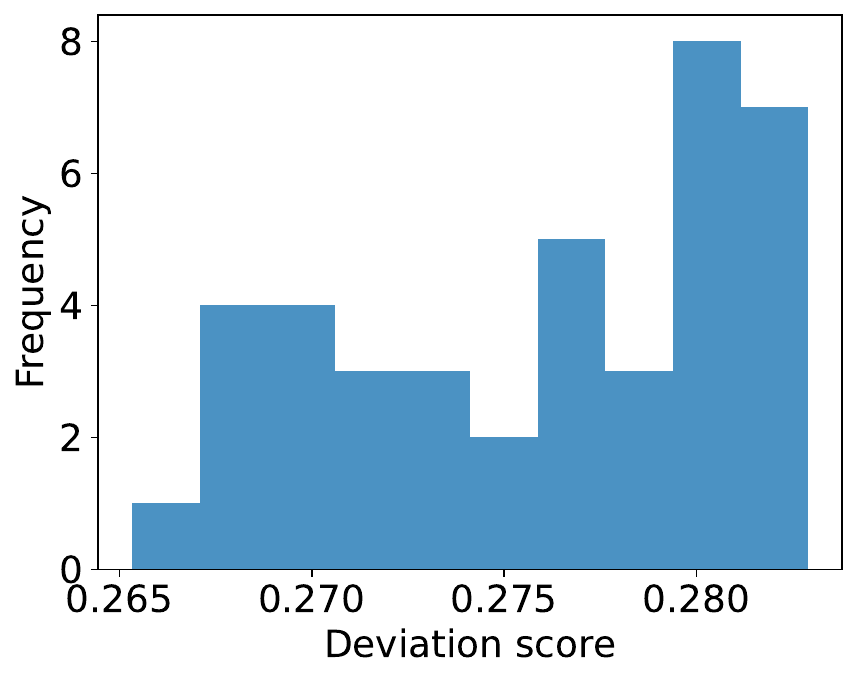}}}
    \hfill
    \subfloat[MMLU/GPT-3.5/CoT]{\label{fig:finqa3}{\includegraphics[width=0.25\textwidth]{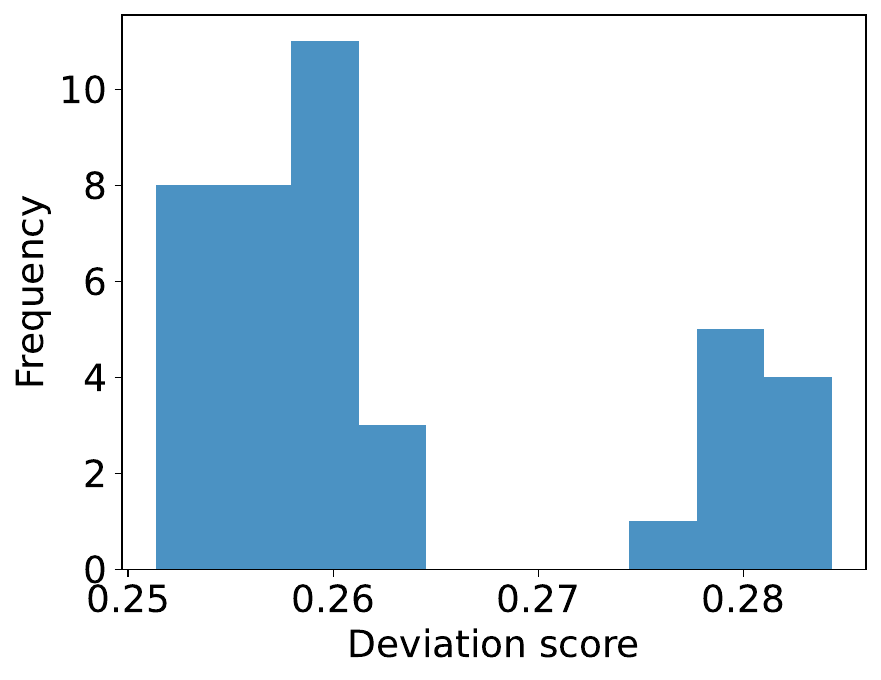}}}
    \hfill
    \subfloat[MMLU/GPT-4o/Vanilla]{\label{fig:finqa3}{\includegraphics[width=0.25\textwidth]{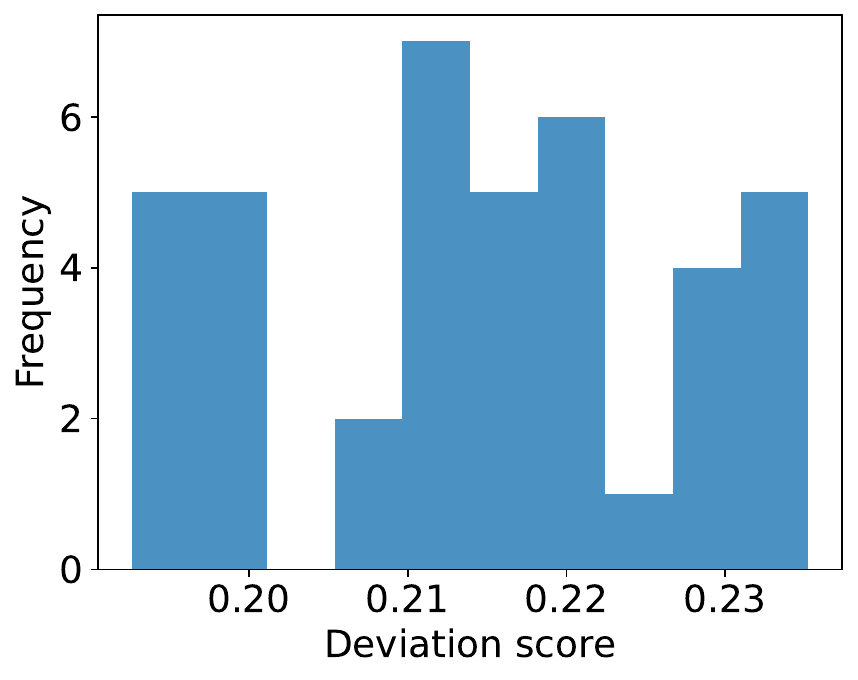}}}
    \hfill
    \subfloat[MMLU/GPT-4o/CoT]{\label{fig:finqa3}{\includegraphics[width=0.25\textwidth]{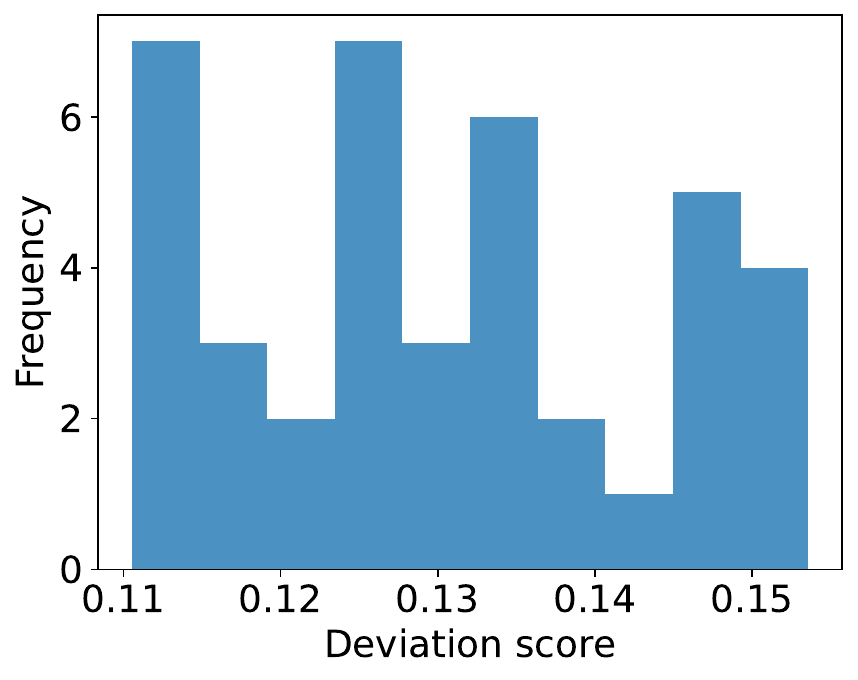}}}
    \caption{{ The figures show the distribution of the average DS metric across confidence levels for different datasets, in different models for vanilla and CoT prompts. GPT-3.5 is short for GPT-3.5-turbo, and GPT-4o is short for GPT-4o-mini. The DS values are lowest for MMLU, higher for the Medical dataset, and highest for FinQA. This supports earlier results in Tables \ref{tab:overprec} and \ref{tab:refine_agg_res_single} and Section \ref{sec:res_eval}, confirming that the observed trends are consistent across datasets and not driven by outliers.}}
    \label{fig:ds_scores}
\end{figure*}

\begin{figure*}
    \centering
    \subfloat[FinQA/GPT-3.5/Vanilla]{\label{fig:finqa3}{\includegraphics[width=0.25\textwidth]{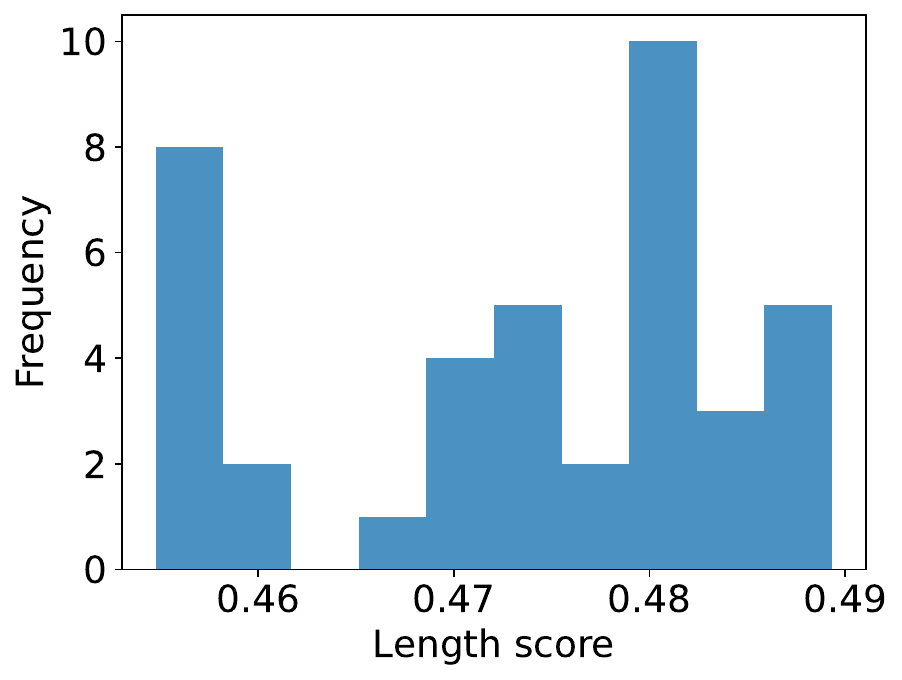}}}
    \hfill
    \subfloat[FinQA/GPT-3.5/CoT]{\label{fig:finqa3}{\includegraphics[width=0.25\textwidth]{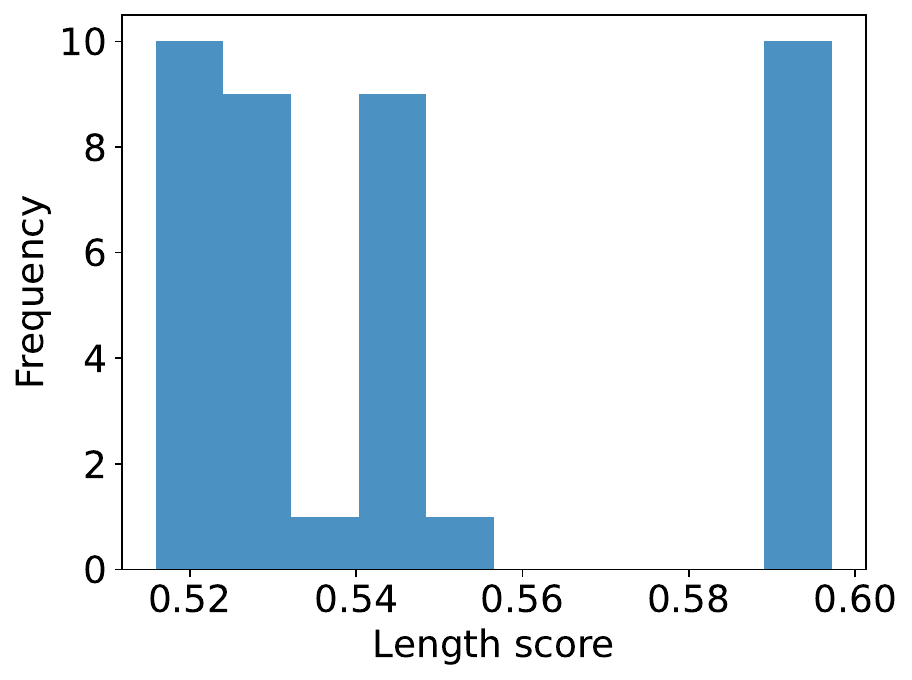}}}
    \hfill
    \subfloat[FinQA/GPT-4o/Vanilla]{\label{fig:finqa3}{\includegraphics[width=0.25\textwidth]{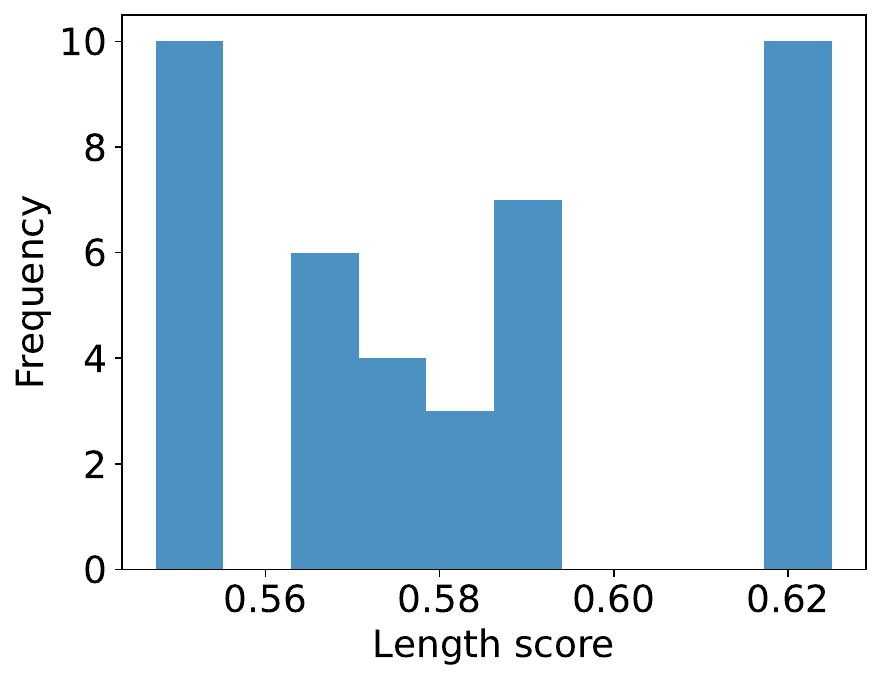}}}
    \hfill
    \subfloat[FinQA/GPT-4o/CoT]{\label{fig:finqa3}{\includegraphics[width=0.25\textwidth]{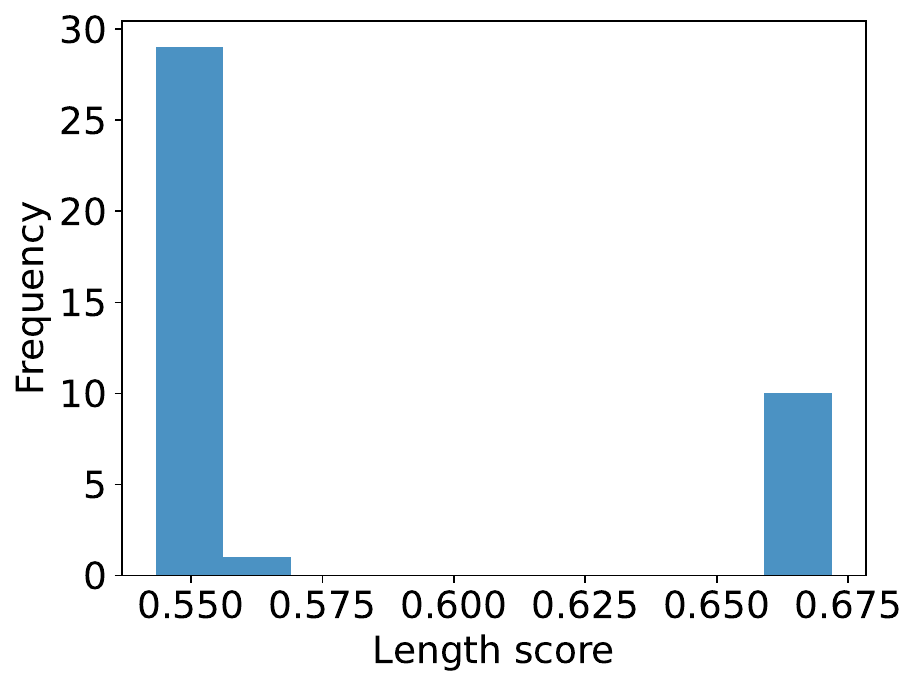}}}
    \hfill
    \subfloat[Medical/GPT-3.5/Vanilla]{\label{fig:finqa3}{\includegraphics[width=0.25\textwidth]{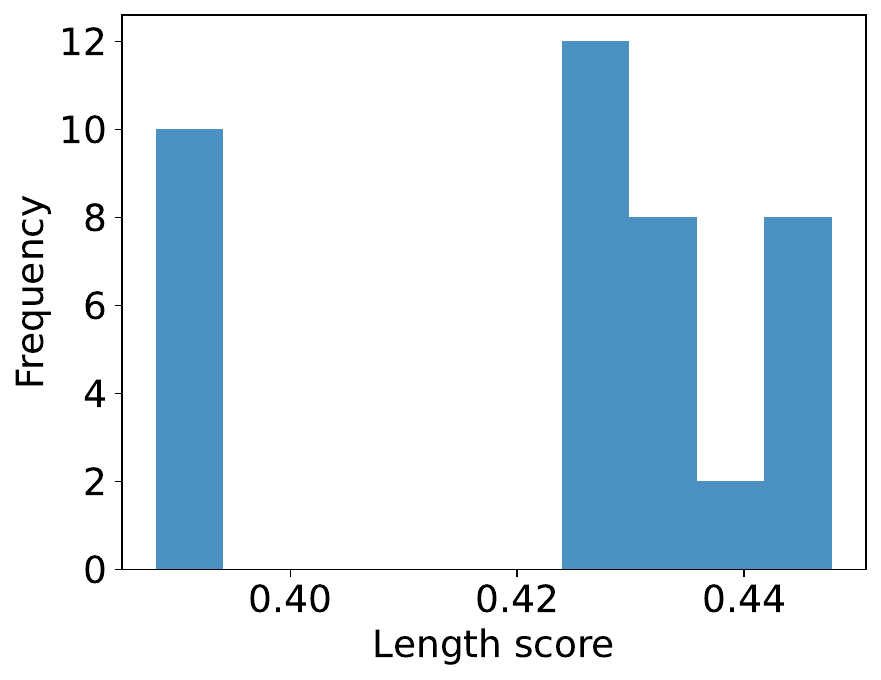}}}
    \hfill
    \subfloat[Medical/GPT-3.5/CoT]{\label{fig:finqa3}{\includegraphics[width=0.25\textwidth]{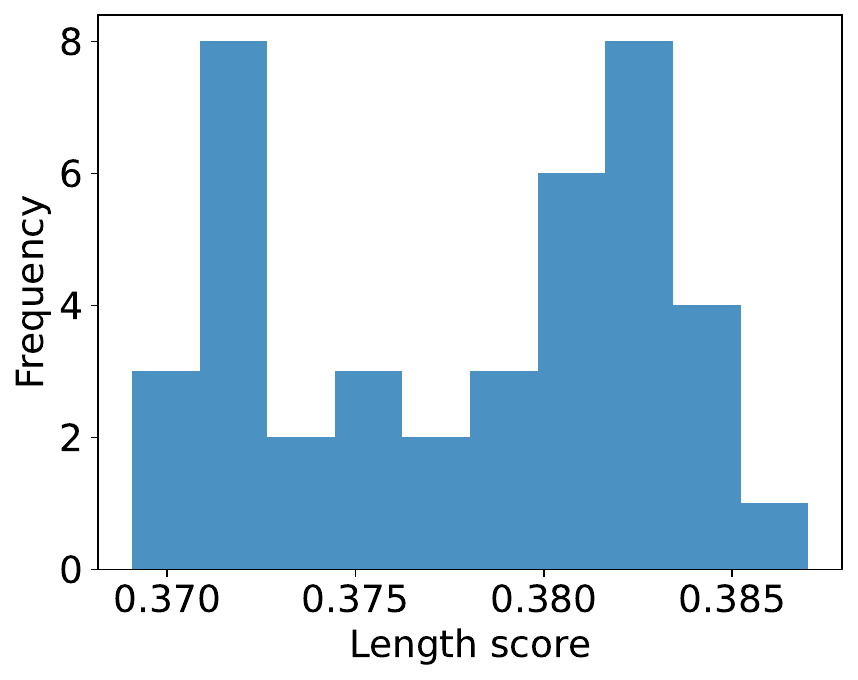}}}
    \hfill
    \subfloat[Medical/GPT-4o/Vanilla]{\label{fig:finqa3}{\includegraphics[width=0.25\textwidth]{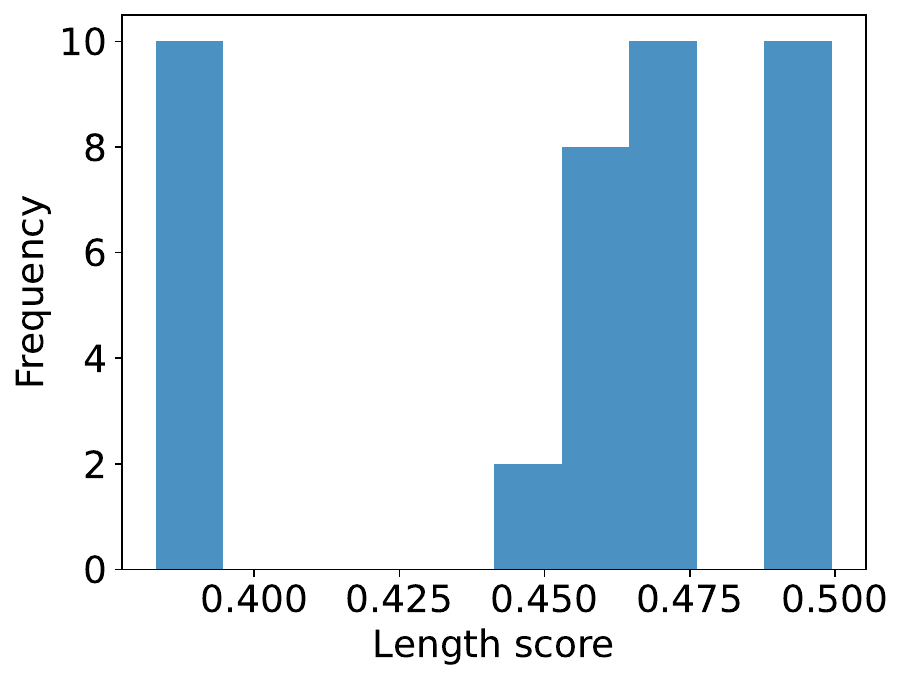}}}
    \hfill
    \subfloat[Medical/GPT-4o/CoT]{\label{fig:finqa3}{\includegraphics[width=0.25\textwidth]{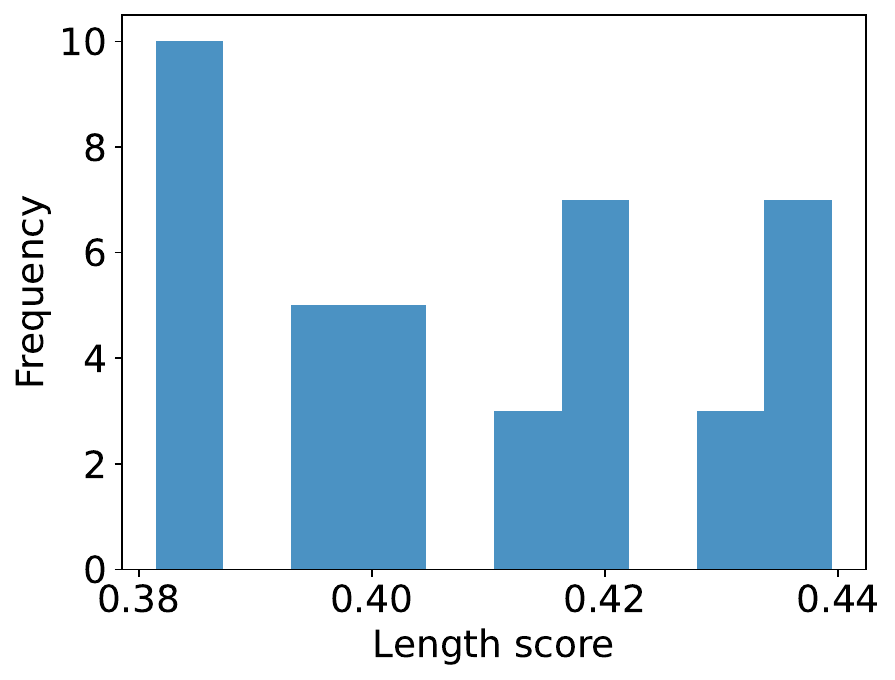}}}
    \hfill
    \subfloat[MMLU/GPT-3.5/Vanilla]{\label{fig:finqa3}{\includegraphics[width=0.25\textwidth]{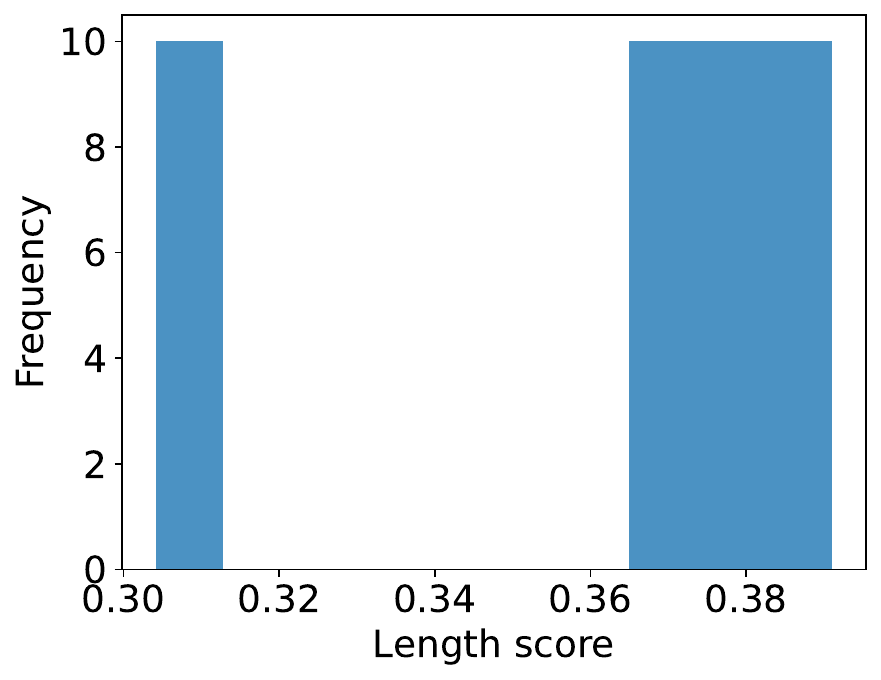}}}
    \hfill
    \subfloat[MMLU/GPT-3.5/CoT]{\label{fig:finqa3}{\includegraphics[width=0.25\textwidth]{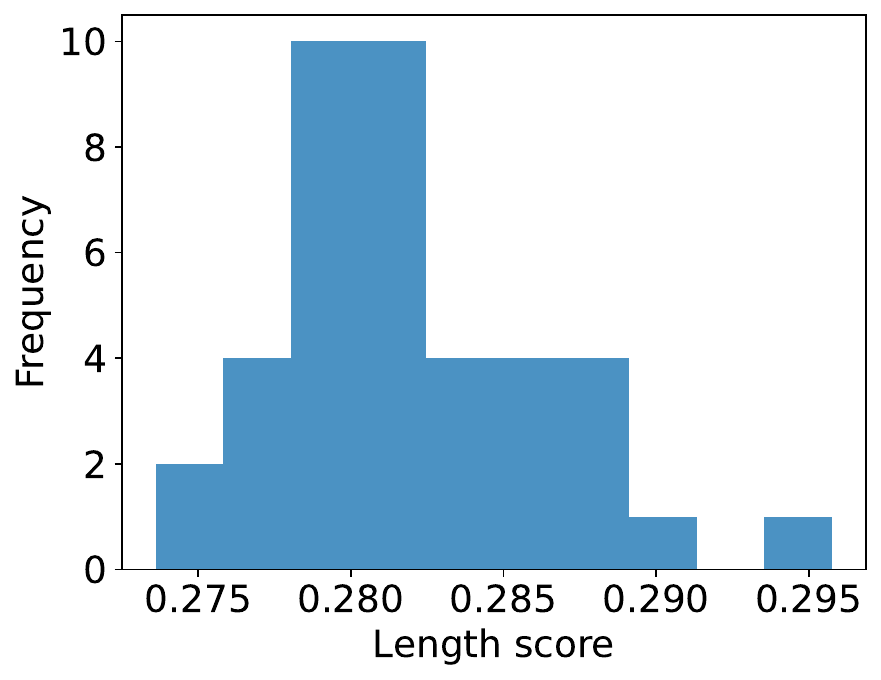}}}
    \hfill
    \subfloat[MMLU/GPT-4o/Vanilla]{\label{fig:finqa3}{\includegraphics[width=0.25\textwidth]{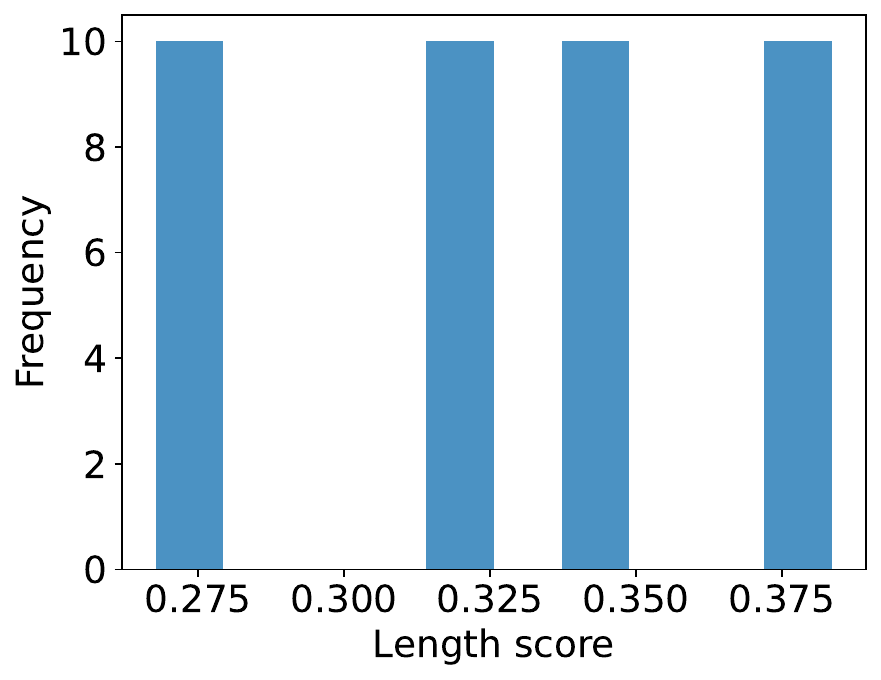}}}
    \hfill
    \subfloat[MMLU/GPT-4o/CoT]{\label{fig:finqa3}{\includegraphics[width=0.25\textwidth]{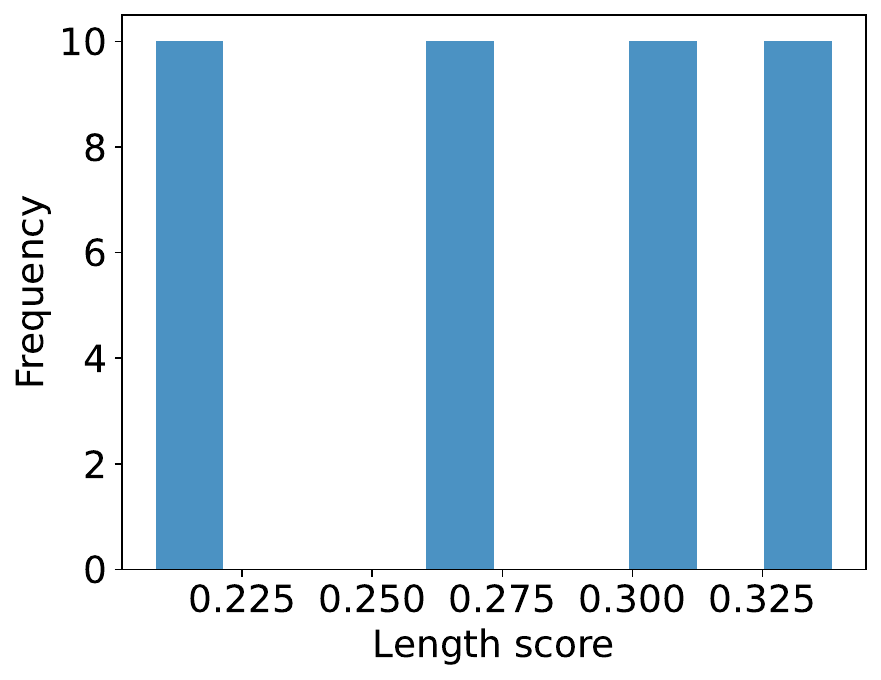}}}
    \caption{{ The figures show the distribution of the average ILS metric across confidence levels for different datasets, in different models for vanilla and CoT prompts. GPT-3.5 is short for GPT-3.5-turbo, and GPT-4o is short for GPT-4o-mini. The figures show that interval lengths are largest in FinQA, followed by Medical, and smallest in MMLU. This suggests that LLMs adjust interval size based on task difficulty, reflecting an awareness of uncertainty. However, combined with earlier findings on the lack of correlation between confidence and interval size, it indicates that while LLMs sense task hardness, they struggle to align their confidence with explicit instructions.}}
    \label{fig:ils_scores}
\end{figure*}

    Figure \ref{fig:selfrevevol} shows how the performance of GPT-4o-mini in the self-refinement process as a function of the number of provided examples for different datasets in different settings. The "chosen" answers performance is not consistent across datasets and settings. However, the accuracy of the proposed responses generally increases with the number of examples in most settings and datasets (except MMLU in the "single" setting). The general trend of improved accuracy with an increasing number of examples suggests that the model benefits from seeing more context or task-specific information during the self-refinement process. This aligns with the principle that additional examples provide more opportunities for the model to learn patterns or clarify ambiguities, especially in few-shot learning setups.
{ 
\section{The effects of different experimental settings on the length and deviation of the intervals}
\label{sec:dev_len_var}
    To study trends and variations in interval size and the deviation from the interval, we introduce two metrics: a) deviation score (DS) and b) interval length score (ILS). The DS measures the amount that the interval deviates from the expected answer, and the ILS measures how large the predicted interval is. DS can be expressed as follows:
    \begin{equation}
       \text{DS}^{c}=\frac{1}{|\hat{A}^c|} \sum_{i=1}^{|\hat{A}^c|}\left(\frac{\max(m_{i}^{c}, 0)}{|m_{i}^{c}|+1}\right)^{2}
   \end{equation}
    with $m_{i}^{c} = \max(x_{i}^{c} - a_{i}, a_{i}-y_{i}^{c})$. This metric equals 0 if the expected answer is in the predicted interval otherwise, the further the answer is from the interval, the higher the score. The ILS metric can be expressed as follows:

    \begin{equation}
        \text{ILS}^{c}=\frac{1}{|\hat{A}^c|} \sum_{i=1}^{|\hat{A}^c|}\frac{y_{i}^{c} - x_{i}^{c}}{\max(|y_{i}^{c}|, |x_{i}^{c}|)}
    \end{equation}
    This metric considers the length of the interval and the scale of the values to penalize larger intervals with lower scales more than smaller intervals with larger scales.
    
    In this section, we study the effects of different datasets, models and prompting techniques on the length and deviation of the intervals. Figures \ref{fig:ds_scores} and \ref{fig:ils_scores} show the distributions of the average DS and ILS metrics for all confidence levels, respectively, in various experimental settings.

    Figure \ref{fig:ds_scores} shows that the deviation scores are lower in MMLU relative to Medical dataset, which in turn has lower scores than those of the FinQA dataset. This reinforces the results shown in tables \ref{tab:overprec} and \ref{tab:refine_agg_res_single} and the findings in section \ref{sec:res_eval}, and demonstrates that those trends are not produced by outliers, but are consistent across each dataset.
    
    Figure \ref{fig:ils_scores} shows that the average lengths of the intervals in FinQA dataset are larger than those of Medical dataset, which also has intervals larger than the MMLU dataset. This demonstrates that an LLM varies its interval size depending on how certain it is of the answer, which in addition to the previous findings about the lack of correlation between the confidence level and interval size, shows that LLMs can't adjust their confidence following instructions but they are still aware at a certain level of the task hardness and their lack of knowledge.

    The effects of the different choices of prompting techniques and LLMs on the ILS and DS are mixed. In some cases, GPT-4o-mini significantly improved the ILS and DS over GPT-3.5-Turbo, and in some cases, the effect of model change is negligible or reversed. The same can be said for prompting techniques.
    }

\end{document}